\documentclass[11pt]{article}

\usepackage[preprint]{acl}

\usepackage{times}
\usepackage{latexsym}
\usepackage{amssymb}
\usepackage{amsmath}

\usepackage[T1]{fontenc}

\usepackage[utf8]{inputenc}

\usepackage{microtype}

\usepackage{inconsolata}

\usepackage{graphicx}

\usepackage{amsmath}
\usepackage{amssymb}
\usepackage{algorithm}
\usepackage[noend]{algpseudocode}
\algnewcommand\Input{\item[\textbf{Input:}]}
\algnewcommand\Output{\item[\textbf{Output:}]}
\usepackage{colortbl} %
\usepackage{booktabs}      %
\usepackage{multirow}
\usepackage{graphicx}      %
\usepackage{xcolor}

\title{\textsc{D$^2$Plan}: Dual-Agent Dynamic Global Planning for \\ Complex Retrieval-Augmented Reasoning}

\author{Kangcheng Luo, Tinglang Wu \and Yansong Feng \\
        Peking University \\ \texttt{\{luokangcheng, wutinglang, fengyansong\}@stu.pku.edu.cn}}

\begin{document}

\maketitle

\begin{abstract}
Recent search-augmented LLMs trained with reinforcement learning (RL) can interleave searching and reasoning for multi-hop reasoning tasks.
However, they face two critical failure modes as the accumulating context becomes flooded with both crucial evidence and irrelevant information: 
(1) ineffective search chain construction that produces incorrect queries or omits retrieval of critical information, and (2) reasoning hijacking by peripheral evidence that causes models to misidentify distractors as valid evidence.
To address these challenges, we propose \textsc{D$^2$Plan}, a \textbf{D}ual-agent \textbf{D}ynamic global \textbf{Plan}ning paradigm for complex retrieval-augmented reasoning.
\textsc{D$^2$Plan} operates through the collaboration of a \textit{Reasoner} and a \textit{Purifier}: 
the \textit{Reasoner} constructs explicit global plans during reasoning and dynamically adapts them based on retrieval feedback; 
the \textit{Purifier} assesses retrieval relevance and condenses key information for the \textit{Reasoner}.
We further introduce a two-stage training framework consisting of supervised fine-tuning (SFT) cold-start on synthesized trajectories and RL with plan-oriented rewards to teach LLMs to master the \textsc{D$^2$Plan} paradigm.
Extensive experiments demonstrate that \textsc{D$^2$Plan} enables more coherent multi-step reasoning and stronger resilience to irrelevant information, thereby achieving superior performance on challenging QA benchmarks.
\end{abstract}

\section{Introduction}
\label{sec:introduction}

Recent advancements in reinforcement learning (RL) have empowered Large Language Models (LLMs) with the capability to perform autonomous retrieval during reasoning~\citep{song2025r1, jin2025search, chen2025learning, sun2025zerosearchincentivizesearchcapability, zheng-etal-2025-stepsearch, zhang-etal-2025-lets}. 
Transcending the traditional Retrieval-Augmented Generation (RAG) paradigm~\citep{lewis2020retrieval, gao2023retrieval}, these models are no longer passive consumers of retrieved documents. 
Instead, they actively determine \textit{when}, \textit{what}, and \textit{how} to search, subsequently integrating the retrieved information into their reasoning processes~\citep{lin2025comprehensive}, thereby enhancing performance on knowledge-intensive multi-hop reasoning tasks.

\begin{figure}[t!]
    \centering
    \includegraphics[width=1\linewidth]{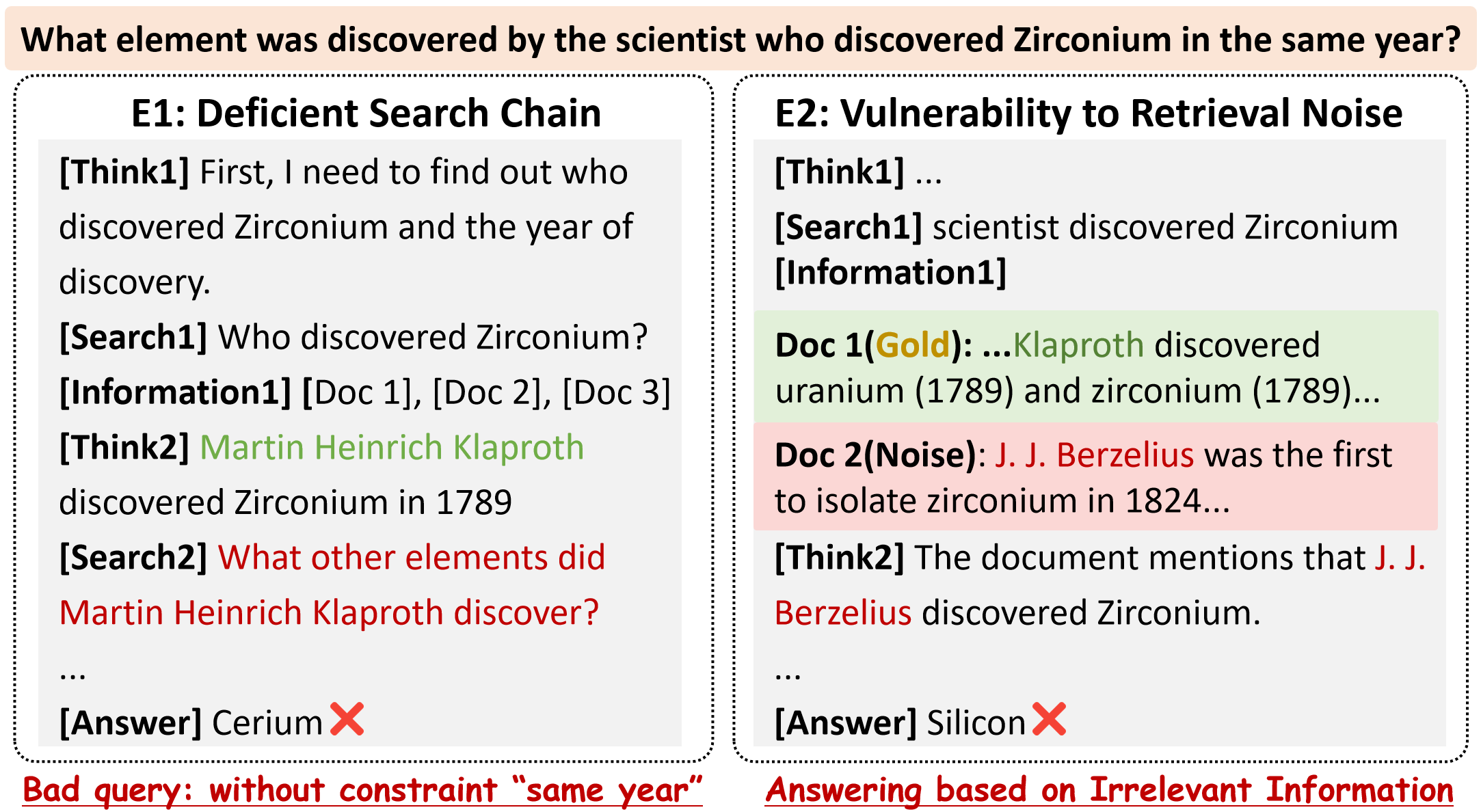}
    \\
    \caption{Illustrative examples of 2 typical error types.}
    \label{fig:error_analysis}
\end{figure}

However, we observe that as LLMs iteratively invoke retrievers to gather evidence for complex questions, the context becomes inundated with documents retrieved at various steps. 
This lengthy context, containing both key evidence and substantial irrelevant information, dilutes the model's attention to the cues in the original question and retrieved critical evidence~\citep{hsieh2024found, fang2025attentionrag}. 
As shown in Figure~\ref{fig:error_analysis} (Left), the model loses track of temporal constraints in the original question, leading to a newly constructed query that lacks global context and causes a drift in retrieval targets. 
In other scenarios (Figure~\ref{fig:error_analysis}, Right), when faced with a lengthy context flooded with all retrieved information, the model struggles to discriminate between valid evidence and distractors, leading to erroneous conclusions derived from irrelevant information. 
We categorize these pervasive failure modes into two primary types:
\textbf{(E1) Ineffective Search Chain Construction:} The query sequences iteratively generated by models often contain incorrect queries, causing divergence from the original reasoning objective, or lack necessary queries, omitting retrieval of essential knowledge points.
\textbf{(E2) Reasoning Hijacking by Peripheral Evidence:} Models tend to misidentify irrelevant information in retrieved results as valid evidence, and fail to re-retrieve when retrieval yields entirely irrelevant results.

After analyzing the results of state-of-the-art methods, we find that both E1 and E2 are prevalent and account for 29\% and 63\% of total failures, respectively (\S\ref{subsec:error_analysis}).
These phenomena suggest that to better deal with complex multi-hop reasoning tasks, models need precise perception and control over the retrieval-augmented reasoning process. Specifically, two core capabilities are essential:
\textbf{(C1) Dynamic Global Planning:} The model must be capable of formulating an explicit global plan and maintaining awareness of it throughout the reasoning process. Furthermore, it should be able to dynamically adapt this plan based on retrieval feedback, including revising sub-questions and rewriting queries.
This ability prevents the drift in retrieval targets and the omission of critical reasoning or retrieval steps, while enabling recovery from retrieval failures.
\textbf{(C2) Robust Resistance to Peripheral Evidence:} The model should be able accurately distinguish key evidence from irrelevant information in retrieved results. 
This capability ensures reasoning is grounded in reliable evidence and enables the model to recognize retrieval failures, thereby performing self-correction by triggering plan adaptation in C1.

To equip models with such capabilities, we propose \textsc{D$^2$Plan}, a \textbf{D}ual-agent \textbf{D}ynamic global \textbf{Plan}ning paradigm supported by a two-stage training framework. 
\textsc{D$^2$Plan} operates through the collaboration of two agents: a \textit{Reasoner} and a \textit{Purifier}. 
The \textit{Reasoner} performs retrieval-augmented reasoning with explicit dynamic global planning, while the \textit{Purifier} evaluates retrieval relevance and distills key information for the \textit{Reasoner}. 
For a given question, the \textit{Reasoner} autonomously constructs a global plan (i.e., an ordered sequence of sub-questions), sequentially addresses each sub-question, and dynamically adapts the plan based on feedback from the \textit{Purifier}.

The training process of \textsc{D$^2$Plan} consists of two phases: SFT Cold-Start and \textsc{SPlanRL}. 
In the first phase, we design a pipeline to manually implement the workflow of \textsc{D$^2$Plan} and synthesize data using a powerful teacher model to fine-tune both the \textit{Reasoner} and \textit{Purifier}. 
This cold-start stage enables the \textit{Reasoner} to acquire preliminary reasoning capabilities with dynamic global planning, and enables the \textit{Purifier} to learn accurate relevance assessment and information condensation. 
In the second phase, we introduce \textsc{SPlanRL}, which utilizes plan-oriented rewards to enhance the \textit{Reasoner}'s reasoning capability by guiding it to correctly formulate and adapt global plans.
To validate the effectiveness of \textsc{D$^2$Plan}, we conducted extensive experiments on challenging QA benchmarks. 
Notably, our method achieves an average absolute performance improvement of 3.8\% over the state-of-the-art baseline across six datasets.

Our contributions are as follows:
(1) We identify two prevalent failure modes of existing LLMs when handling lengthy retrieval-augmented reasoning contexts: Ineffective Search Chain Construction and Reasoning Hijacking by Peripheral Evidence.
(2) We propose \textsc{D$^2$Plan}, a dual-agent dynamic planning paradigm supported by a two-stage training framework, to enhance the global planning and resistance to peripheral evidence of search-augmented LLMs, thereby mitigating these failure modes.
(3) Extensive experiments demonstrate that \textsc{D$^2$Plan} enables more coherent query formulation and stronger resilience to peripheral evidence during reasoning, achieving superior performance on challenging QA benchmarks.

\section{Methodology}
\begin{figure*}[t!]
    \centering
    \includegraphics[width=0.9\linewidth]{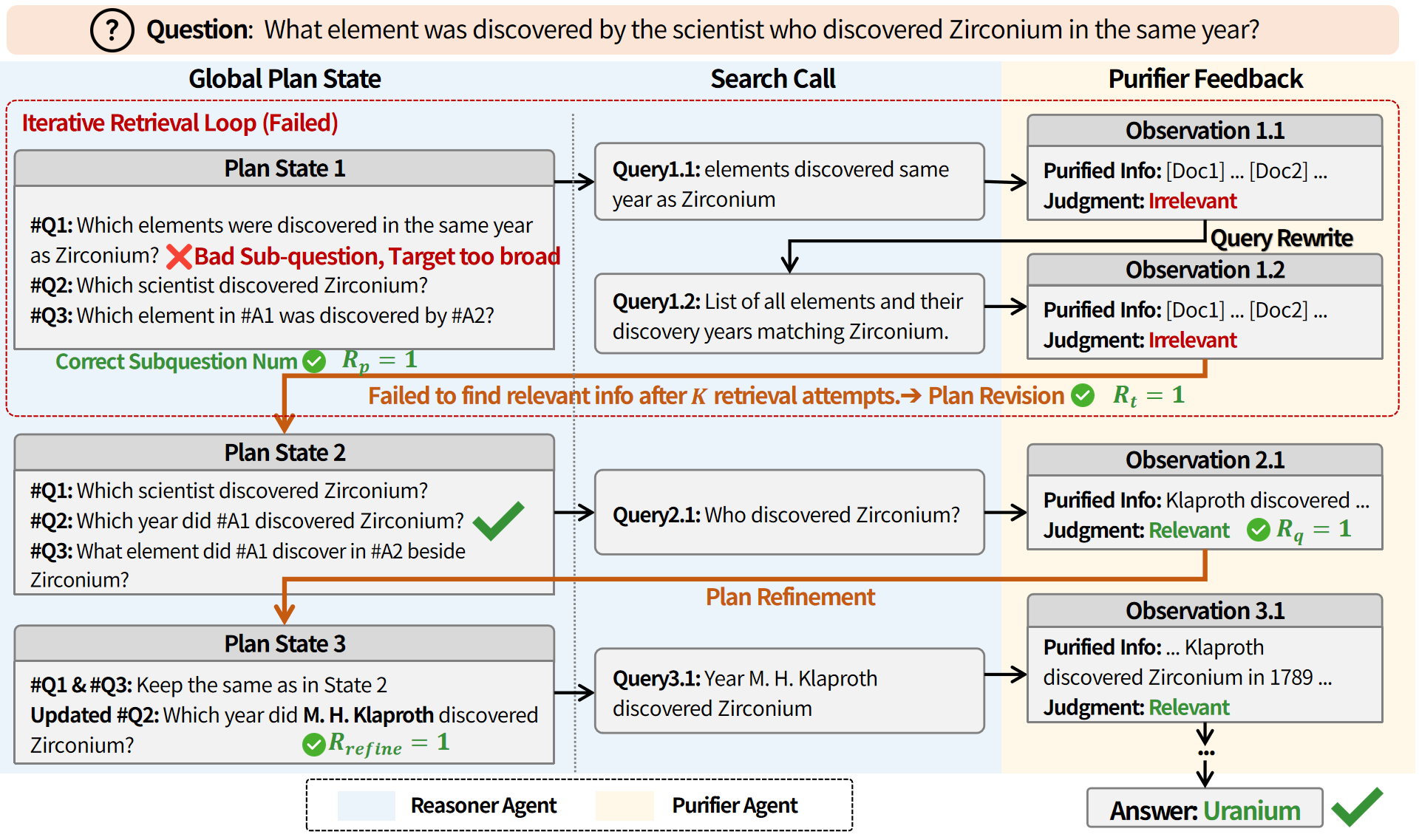}
    \caption{An illustrative example of our \textsc{D$^2$Plan} paradigm resolving a multi-hop reasoning question, where the correct number of sub-questions in the initial plan and effective plan adaptation are rewarded in the RL phase.}
    \label{fig:main_fig}
\end{figure*}
In this section, we present our \textsc{D$^2$Plan} paradigm and the corresponding training framework. 
First, we provide an overview of the \textsc{D$^2$Plan} workflow\footnote{See Appendix~\ref{app:workflow} for pseudocode of the workflow.}, introducing how its dual-agent architecture and dynamic global planning mechanism operate when solving challenging problems (\S\ref{subsec:overview}).
Then, we detail how we teach LLMs to master this paradigm through a two-stage training framework comprising SFT cold-start (\S\ref{subsec:sft}) and \textsc{SPlanRL} (\S\ref{subsec:rl}).
\subsection{Overview of \texorpdfstring{\textsc{D$^2$Plan}}{D2Plan}}
\label{subsec:overview}
\begin{figure}[t]
    \centering
    \includegraphics[width=\linewidth]{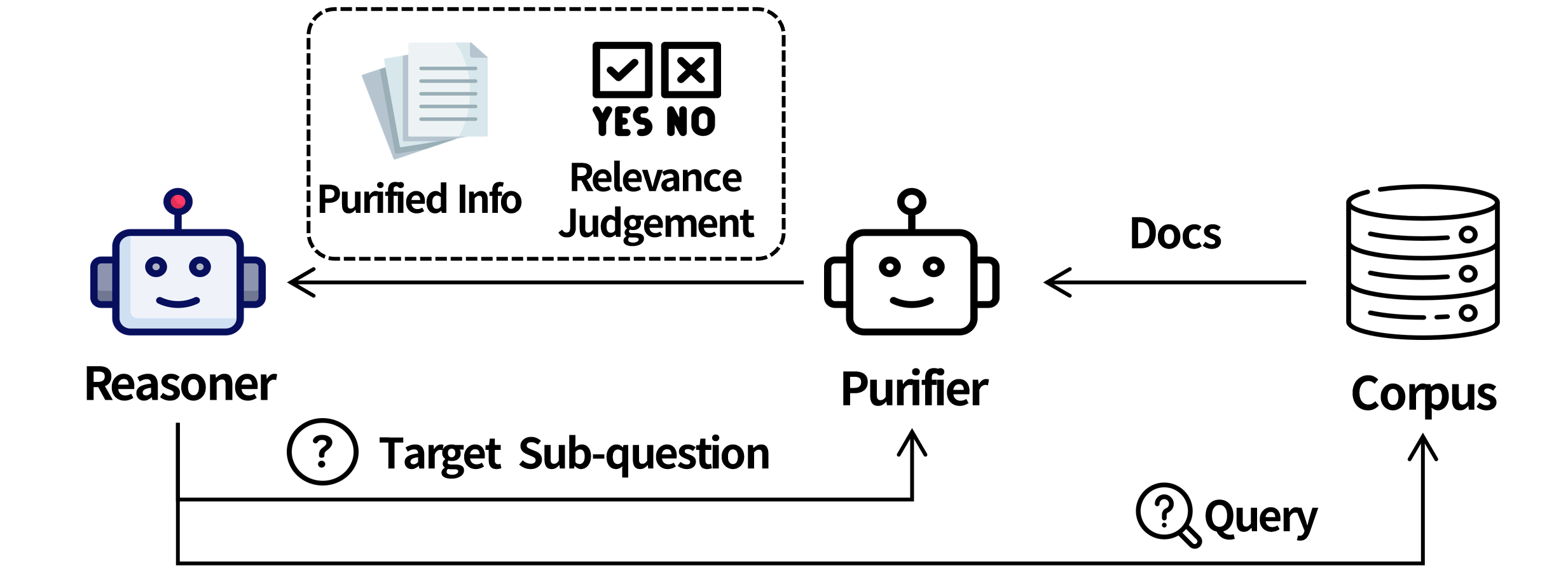}
    \caption{Overview of the dual-agent architecture.
     } %
    \label{fig:dual_agent}
\end{figure}

\noindent \textbf{Dual-Agent Architecture} \quad
As shown in Figure~\ref{fig:dual_agent}, \textsc{D$^2$Plan} comprises two agents: a \textit{Reasoner} and a \textit{Purifier}.
The \textit{Reasoner} is responsible for executing core planning, reasoning, and invoking search tools. 
When invoking search tools, the \textit{Reasoner} generates a query and specifies the target sub-question to be solved.
A retriever returns candidate documents based on the query. These documents are not directly passed to the \textit{Reasoner}, but first analyzed by the \textit{Purifier}.
The \textit{Purifier} focuses on assessing whether the candidate documents contain key information for solving the target sub-question.
If relevant information exists, the \textit{Purifier} extracts it precisely; otherwise, it provides a brief summary of the retrieved documents.
The purified information and relevance judgment are then fed back to the \textit{Reasoner}, enabling it to adapt the global plan and continue reasoning.

\noindent \textbf{Dynamic Global Planning} \quad
As shown in Figure~\ref{fig:main_fig}, given an input question, the \textit{Reasoner} autonomously assesses its complexity and constructs a global problem-solving plan.
Single-hop questions require no decomposition, whereas multi-hop questions are decomposed into an ordered sequence of sub-questions to form an initial plan.
In this sequence, each sub-question depends only on preceding ones and can reference their answers using placeholders like \texttt{\#A\_j} (denoting the answer to the $j$-th sub-question), which allows the \textit{Reasoner} to solve them sequentially.
When solving each sub-question, the \textit{Reasoner} can iteratively generate queries and invoke search tools based on feedback from the \textit{Purifier}, continuing until relevant evidence is obtained or the maximum number of attempts $K$ is reached.
Upon successfully solving a sub-question, the \textit{Reasoner} refines the subsequent sub-question based on the resolved context, making it more specific and self-contained to facilitate effective retrieval and
prevent divergence from the original reasoning objective. 
Conversely, if a sub-question remains unsolved after reaching the maximum retrieval attempts, the \textit{Reasoner} revises the remaining sub-questions, including the current unsolved one, to explore alternative problem-solving paths for self-correction.

\subsection{SFT Cold-Start}
\label{subsec:sft}
We use SFT to train the dual agents, guiding \textit{Reasoner} to acquire preliminary retrieval-augmented reasoning capabilities with dynamic global planning, and enabling \textit{Purifier} to master relevance assessment and information condensation.
\subsubsection{Synthesizing Trajectories}
\label{subsec:synthesising_trajectories}
To obtain high-quality SFT training data, we design a pipeline that guides a powerful teacher LLM to play the roles of \textit{Reasoner} and \textit{Purifier}.\footnote{See Appendix~\ref{app:trajectory_synthesis} for detailed algorithms and pseudocode.}
Specifically, we employ a series of curated prompts (see Appendix~\ref{app:prompts}) to step-by-step guide the teacher model to simulate the workflow described in \S\ref{subsec:overview}, executing a series of operations including question decomposition, query rewriting, plan refinement, and plan revision, etc.
Through this process, we obtain a set of reasoning trajectories synthesized according to the \textsc{D$^2$Plan} workflow.
Each trajectory is also accompanied by the corresponding input-output pairs of the \textit{Purifier} for relevance judgment and information purification at each retrieval step.

\subsubsection{Constructing SFT Data}
\label{subsubsec:constructing_sft_data}
To construct high-quality SFT training data, we first filter the synthesized reasoning trajectories based on two criteria: 
(1) the final answer exactly matches the gold answer; 
(2) the trajectory contains at least one retrieval call. 
The second criterion ensures that the model prioritizes learning to leverage retrieval tools for problem-solving rather than relying on potentially erroneous internal parametric knowledge.
Based on this filtered set of high-quality trajectories, we construct corresponding training data for both agents.
For the \textit{Reasoner}, we further use the teacher LLM to reformat the filtered reasoning trajectories, integrating the segmented reasoning steps synthesized via multiple prompts into a coherent and fluent reasoning chain that serves as the training target.
For the \textit{Purifier}, we extract the input-output pairs corresponding to each retrieval step from the filtered reasoning trajectories as its training data.

\subsection{\textsc{SPlanRL}}
\label{subsec:rl}

After the SFT stage, we freeze the \textit{Purifier} and introduce \textsc{SPlanRL} to further enhance the \textit{Reasoner}'s reasoning capability by strengthening its dynamic planning proficiency.
In addition to the commonly used format and answer rewards in general RL, we design plan-oriented rewards to effectively guide the model's dynamic planning behavior, incentivizing both the quality of plan construction and the effectiveness of plan adjustments.

\noindent \textbf{Initial Planning Reward} \quad
This reward incentivizes the model to identify the required number of reasoning steps and construct high-quality initial plans.
Specifically, for single-hop questions, we assign $R_{p}=1$ if the model does not perform question decomposition, and $R_{p}=0$ otherwise.
For multi-hop questions, we assign $R_{p}=1$ if the model performs question decomposition and the number of generated sub-questions matches the hop count annotated in the dataset, and $R_{p}=0$ otherwise.
We use count-based rather than semantic constraints for two reasons.
First, count-based constraints do not require annotations of sub-questions and sub-answers, making them more practical to implement.
Second, count-based constraints effectively prevent the model from generating redundant sub-questions or missing necessary ones, thereby improving the conciseness and effectiveness of planning (\S\ref{para:rp_analysis}).

\noindent \textbf{Plan Adaptation Reward} \quad
This reward aims to reinforce the model's capability of dynamically adjusting plans throughout the reasoning process, encompassing two complementary mechanisms: \textit{plan refinement} and \textit{plan revision}. 

\textit{Plan Refinement Reward.} %
For each sub-question that depends on the preceding ones, we use the Plan Refinement Reward $R_{refine}$ to encourage the model to refine it after solving those dependencies---filling in placeholders and making subsequent sub-questions more specific and self-contained.
A successful refinement requires that the refined sub-question exists in the reasoning trajectory and contains no placeholders, which we verify using regular expression matching.
If all such sub-questions are successfully refined, the reward $R_{refine}=1$, and $R_{refine}=0$ otherwise.
This reward not only encourages the model to improve sub-question quality, but also effectively prevents the drift in retrieval targets during multi-hop reasoning.

\textit{Plan Revision Reward.} We design a revision reward to encourage the model to globally revise the remaining sub-questions when iterative retrieval reaches the maximum number of attempts without finding relevant information.
Such failures are often caused by poorly formulated sub-questions, such as those that are overly vague or lack necessary constraints.
This reward is jointly determined by timing reward $R_{t}$ and quality reward $R_{q}$.
$R_{t}=1$ if the model triggers revision only after retrieval reaches the maximum number of failures, preventing premature abandonment; otherwise $R_{t}=0$. 
$R_{q}=1$ if the first revised sub-question (i.e., the current unsolved sub-question after revision) successfully retrieves relevant information; otherwise $R_{q}=0$. 
We use the relevance judgment from the \textit{Purifier} to determine whether the retrieval is successful.
Specifically, let $c \in \{0, 1\}$ denote answer correctness and $r \in \{0, 1\}$ indicate whether revision was triggered, then:
\begin{equation}
\resizebox{0.8\linewidth}{!}{$
R_{revise} = 
\begin{cases} 
1 & \text{if } c = 1 \\ 
\lambda \cdot (R_{t} + R_{q}) & \text{if } c = 0 \land r = 1 \\ 
0 & \text{otherwise} 
\end{cases}
$}
\end{equation} %
By introducing a discount factor $\lambda$, $R_{revise}$ provides partial process rewards for trajectories that answer incorrectly but attempt revision.
This design effectively guides the model to revise its plan in the right direction and acquire the "failure-reflection-correction" reasoning pattern.

Combining the two terms, our Plan Adaptation Reward is defined as $R_{a} = R_{refine} + R_{revise}$.

\noindent \textbf{Answer Reward} \quad
Following recent work~\citep{zheng-etal-2025-stepsearch,zhang-etal-2025-lets}, we adopt the word-level F1 score between the predicted answer and the ground truth to measure answer correctness.
The answer reward is defined as $R_{ans} = \text{F1}(a_{pred}, a_{gt})$, where $\text{F1}(a_{pred}, a_{gt})$ denotes the F1 score between the predicted answer $a_{pred}$ and the golden answer $a_{gt}$.

\noindent \textbf{Format Reward} \quad
We treat the format reward as a prerequisite constraint, aiming to ensure the model to generate parsable reasoning paths, which is essential for computing the Initial Planning Reward $R_p$ and Plan Adaptation Reward $R_a$.
For example, the model must wrap the initial plan with \texttt{<plan></plan>} tags, the revised plan with \texttt{<replan></replan>} tags, and the final answer with \texttt{<answer></answer>} tags (see Appendix~\ref{app:format_constraints} for detailed constraints).
When all constraints are satisfied, $R_{f}=1$, otherwise 0.

\noindent \textbf{Total Reward} \quad
The final total reward function is defined as:
        $
R_{total} = R_{f} \cdot (\alpha R_{p} + \beta R_{a} + R_{ans}), 
    $
where $\alpha$ and $\beta$ are scaling factors that control the contribution of each reward component.

\section{Experimental Setup}
\noindent \textbf{Datasets and Evaluation Metrics} \quad
To comprehensively evaluate the capabilities of \textsc{D$^2$Plan}, we conduct experiments on widely used open-domain QA benchmarks, including Natural Questions (NQ)~\citep{kwiatkowski2019natural}, SimpleQA~\citep{wei2024measuringshortformfactualitylarge}, HotpotQA~\citep{yang2018hotpotqa}, 2WikiMultihopQA (2Wiki)~\citep{ho2020constructing}, MuSiQue (MSQ)~\citep{trivedi2022musique}, and FRAMES~\citep{krishna2025fact}. 
For evaluation metrics, we employ Exact Match (EM) and LLM-as-a-Judge (LasJ). 
For LasJ, we utilize \texttt{qwen-plus}~\citep{qwen2025qwen25technicalreport} to assess the correctness of predicted answers (see Appendix~\ref{app:prompts} for the prompt).

\noindent \textbf{Baselines} \quad
We compare \textsc{D$^2$Plan} with representative RL-based methods: Search-R1~\citep{jin2025search}, ReSearch~\citep{chen2025learning}, AutoRefine~\citep{shi2025search}, ZeroSearch~\citep{sun2025zerosearchincentivizesearchcapability}, R1-Searcher~\citep{song2025r1}, and StepSearch~\citep{zheng-etal-2025-stepsearch}.
We reproduce the inference results of all baseline methods under the same environment to ensure fair comparison.

\noindent \textbf{Implementation Details} \quad
We employ Qwen2.5-3B/7B-Instruct~\citep{qwen2025qwen25technicalreport} as the backbone models.
For retrieval, we utilize E5~\citep{wang2022text} as the embedding model and the 2018 Wikipedia dump~\citep{karpukhin2020dense} as the retrieval corpus, retrieving the top-5 passages.
For SFT cold-start, we use \texttt{qwen-max-latest} as the teacher model to synthesize 4,000 training instances for the \textit{Reasoner} and 11,255 for the \textit{Purifier} from the training sets of NQ, HotpotQA, and MuSiQue.
For RL training, we sample 29,762 instances from the same training sets and adopt GRPO~\citep{shao2024deepseekmath} as the learning algorithm, implemented on verl~\citep{sheng2025hybridflow}.
More implementation details are provided in Appendix~\ref{app:training_details}.

\section{Results and Analysis}
\label{sec:results_analysis}

In this section, we aim to answer the following Research Questions (RQs):
\textbf{RQ1}: How effectively does \textsc{D$^2$Plan} and each of its components enhance performance in retrieval-augmented question answering? (\S\ref{subsec:main_results}, \S\ref{subsec:ablation_study})
\textbf{RQ2}: Can Dynamic Global Planning improve the quality of query formulation? (\S\ref{subsec:retrieval_quality_analysis})
\textbf{RQ3}: Can \textsc{SPlanRL} enable LLMs to generate and adapt higher-quality plans? (\S\ref{subsec:splanrl_analysis})
\textbf{RQ4}: Can Dual-Agent Architecture enhance accuracy and efficiency? (\S\ref{subsec:dual_agent_analysis})
\textbf{RQ5}: Can \textsc{D$^2$Plan} reduce these errors: (E1) Ineffective Search Chain Construction and (E2) Reasoning Hijacking by Peripheral Evidence? (\S\ref{subsec:error_analysis}) 

\subsection{Main Results (RQ1)}
\label{subsec:main_results}
\begin{table*}[h!]
\centering

\resizebox{\textwidth}{!}{%
\begin{tabular}{lcccccccccccccc}
\toprule
\multirow{2}{*}{\textbf{Method}} & \multicolumn{2}{c}{\textbf{NQ}$^\dagger$} & \multicolumn{2}{c}{\textbf{SimpleQA}$^\ddagger$} & \multicolumn{2}{c}{\textbf{HotpotQA}$^\dagger$} & \multicolumn{2}{c}{\textbf{2Wiki}$^\ddagger$} & \multicolumn{2}{c}{\textbf{MuSiQue}$^\dagger$} & \multicolumn{2}{c}{\textbf{FRAMES}$^\ddagger$} & \multicolumn{2}{c}{\textbf{AVG}} \\
\cmidrule(lr){2-3} \cmidrule(lr){4-5} \cmidrule(lr){6-7} \cmidrule(lr){8-9} \cmidrule(lr){10-11} \cmidrule(lr){12-13} \cmidrule(lr){14-15}
 & LasJ & EM & LasJ & EM & LasJ & EM & LasJ & EM & LasJ & EM & LasJ & EM & LasJ & EM \\
\midrule
\rowcolor{gray!20}
	\textit{Base LLM} & \multicolumn{14}{c}{\textit{Qwen2.5-3B-Instruct}} \\
\midrule
AutoRefine & \textbf{57.01} & \underline{41.05} & \textbf{36.20} & \underline{24.90} & \underline{57.41} & \underline{41.12} & \underline{48.08} & \underline{38.05} & \underline{28.26} & \underline{19.49} & \underline{23.79} & \underline{11.17} & \underline{41.79} & \underline{29.30} \\
Search-R1 & 44.21 & 28.70 & 17.70 & 8.20 & 33.49 & 18.07 & 33.06 & 20.92 & 16.01 & 6.25 & 18.20 & 4.85 & 27.11 & 14.50 \\
StepSearch & 45.60 & 32.91 & 28.20 & 20.40 & 37.31 & 26.64 & 33.33 & 28.78 & 13.24 & 8.52 & 13.83 & 6.55 & 28.59 & 20.63 \\
ZeroSearch & 51.36 & 38.42 & 31.50 & 22.80 & 40.61 & 29.44 & 30.51 & 26.42 & 10.30 & 6.00 & 10.32 & 4.00 & 29.10 & 21.18 \\
\midrule
\textbf{\textsc{D$^2$Plan}-3B} & \underline{55.60} & \textbf{41.30} & \underline{34.10} & \textbf{25.50} & \textbf{57.93} & \textbf{43.58} & \textbf{52.53} & \textbf{45.90} & \textbf{33.14} & \textbf{23.13} & \textbf{25.97} & \textbf{14.68} & \textbf{43.13} & \textbf{32.35} \\

\midrule
\rowcolor{gray!20}
	\textit{Base LLM} & \multicolumn{14}{c}{\textit{Qwen2.5-7B-Instruct}} \\
\midrule
R1-Searcher & 53.43 & 40.00 & 34.60 & 25.60 & \underline{59.03} & \underline{44.32} & \underline{52.04} & \underline{46.13} & \underline{29.83} & \underline{21.93} & \underline{25.61} & \underline{14.32} & \underline{42.38} & \underline{32.05} \\
ReSearch & 53.85 & 39.81 & \underline{36.20} & \underline{27.30} & 57.62 & 42.50 & 47.34 & 41.01 & \underline{29.83} & 21.43 & 24.27 & \underline{14.32} & 41.52 & 31.06 \\
Search-R1 & \underline{55.96} & \textbf{42.11} & 34.40 & 23.10 & 51.67 & 37.31 & 39.36 & 33.13 & 21.51 & 15.27 & 19.78 & 10.19 & 37.11 & 26.85 \\
StepSearch & 53.07 & 38.64 & 34.80 & 24.90 & 52.78 & 38.93 & 49.57 & 42.89 & 28.55 & 20.31 & 23.18 & 13.23 & 40.32 & 29.82 \\
ZeroSearch & 54.02 & \underline{41.00} & 33.10 & 24.10 & 45.25 & 33.48 & 36.70 & 31.76 & 14.23 & 9.39 & 18.33 & 10.07 & 33.61 & 24.97 \\
\midrule
\textbf{\textsc{D$^2$Plan}-7B} & \textbf{58.03} & 39.78 & \textbf{36.50} & \textbf{27.50} & \textbf{61.59} & \textbf{46.17} & \textbf{56.32} & \textbf{50.02} & \textbf{33.47} & \textbf{24.41} & \textbf{31.31} & \textbf{18.08} & \textbf{46.21} & \textbf{34.33} \\
\bottomrule
\end{tabular}%
}
\caption{Main results. $^\dagger$/$^\ddagger$ represents in-domain/out-domain datasets.}
\label{tab:main_results}
\end{table*}

As shown in Table~\ref{tab:main_results}, \textbf{\textsc{D$^2$Plan} consistently outperforms all baselines at both 3B and 7B scales}, demonstrating strong performance and generalization across in-domain and out-of-domain benchmarks.
On average, \textsc{D$^2$Plan}-3B achieves a 1.34\% LasJ gain over the strongest baseline AutoRefine, while \textsc{D$^2$Plan}-7B yields a 3.83\% LasJ improvement over strongest baseline R1-Searcher.

Furthermore, \textbf{\textsc{D$^2$Plan}'s advantage widens as reasoning complexity increases}, with relative improvements over Search-R1 scaling from 19.2\% on the simpler HotpotQA to 58.3\% on the more challenging FRAMES benchmark. 
These results demonstrate the efficacy of our dual-agent dynamic global planning paradigm in managing complex retrieval-augmented reasoning process.

\subsection{Ablation Study (RQ1)}
\label{subsec:ablation_study}

\begin{table}[t]
\centering
\resizebox{0.85\linewidth}{!}{
\begin{tabular}{lccc}
\toprule
\textbf{Method} & \textbf{2Wiki} & \textbf{MSQ} & \textbf{FRAMES} \\
\midrule
\rowcolor{gray!20}
  \multicolumn{4}{c}{\textit{Only SFT}} \\
\textsc{D$^2$Plan}-3B & 48.64 & 26.23 & 25.00 \\
\quad w/o Plan\&Adapt & 41.40 & 22.37 & 18.45 \\
\quad w/o Adapt & 46.40 & 23.42 & 21.24 \\
\midrule
\rowcolor{gray!20}
  \multicolumn{4}{c}{\textit{SFT+RL}} \\
\textsc{D$^2$Plan}-3B & 52.53 & 33.14 & 25.97 \\
\quad w/o Plan\&Adapt & 49.98 & 28.13 & 21.84 \\
\quad w/o Adapt & 50.37 & 29.38 & 23.30 \\
\quad w/o $R_{p}$ & 50.59 & 30.45 & 24.88 \\
\quad w/o $R_{a}$ & 51.65 & 30.16 & 24.76 \\
\bottomrule
\end{tabular}
}
\caption{Ablation study of \textsc{D$^2$Plan}. We report LasJ.}
\label{tab:ablation_study}
\end{table}     

To investigate the contributions of each core component to model performance, we conduct ablation studies on \textsc{D$^2$Plan}-3B and report the results in Table~\ref{tab:ablation_study}. 
Here, \textit{Plan} refers to the initial global planning mechanism for decomposing the question into sub-questions, and \textit{Adapt} refers to the plan adaptation mechanism for plan refinement and revision. Detailed implementations of each ablation variant are provided in Appendix~\ref{app:ablation_details}.

Results show that under the \textit{Only SFT} setting, removing the planning components (\textit{w/o Plan\&Adapt} or \textit{w/o Adapt}) significantly degrades performance compared to \textsc{D$^2$Plan}. 
Despite being trained on reasoning trajectories synthesized from the same question set, the lack of dynamic planning mechanism prevents these variants from effectively organizing reasoning paths for complex multi-hop problems.
This confirms that \textsc{D$^2$Plan}'s gains stem from learned dynamic planning capabilities rather than mere knowledge distillation from the teacher model.
Comparing \textit{SFT+RL} with \textit{Only SFT}, RL consistently improves all variants; however, even with RL optimization, \textit{w/o Plan\&Adapt} still lags behind \textsc{D$^2$Plan}, demonstrating that the dynamic planning paradigm provides irreplaceable advantages for structuring complex multi-hop reasoning.
Furthermore, the performance drop caused by excluding the initial planning reward $R_p$ (\textit{w/o $R_p$}) or the plan adaptation reward $R_a$ (\textit{w/o $R_a$}) underscores the necessity of these rewards in guiding the model to internalize robust plan construction and adaptation capabilities.

\subsection{Retrieval Quality (RQ2)}
\label{subsec:retrieval_quality_analysis}

\begin{table}[t]
\centering
\resizebox{0.85\linewidth}{!}{
\begin{tabular}{lcccc}
\toprule
\textbf{Method} & \textbf{VRC} & \textbf{IRC} & \textbf{TRC} & \textbf{SAC} \\
\midrule
\rowcolor{gray!20}
  \multicolumn{5}{c}{\textit{Only SFT}} \\
\textsc{D$^2$Plan}-3B & 2.48 & 1.98 & 4.45 & 1.44 \\
\quad w/o Plan\&Adapt & 1.26 & 1.02 & 2.28 & 1.11 \\
\quad w/o Adapt & 1.57 & 1.08 & 2.65 & 1.22 \\

\bottomrule
\end{tabular}
}
\caption{Analysis of retrieval calls on MuSiQue. VRC: Valid Retrieval Calls; IRC: Invalid Retrieval Calls; TRC: Total Retrieval Calls; SAC: Sub-Answer Coverage.}
\label{tab:retrieval_calls}
\end{table}

To assess the quality of model-generated queries, we count valid and invalid retrieval calls on MuSiQue, as it provides annotated sub-questions and corresponding sub-answers. 
Specifically, we define a \textit{valid retrieval} as a retrieval call where the returned results contain the sub-answer to any of the sub-questions.
We also report \textit{sub-answer coverage}, which measures the average number of sub-answers covered by all retrieval results.
As shown in Table \ref{tab:retrieval_calls}, \textit{w/o Adapt} yields more valid retrieval calls than \textit{w/o Plan\&Adapt} while maintaining a comparable number of invalid retrieval calls, demonstrating that 
\textbf{global planning assists the model in generating higher-quality query sequences by providing explicit sub-question decomposition that effectively guides each retrieval step.}
Furthermore, although \textsc{D$^2$Plan} incurs more retrieval calls than \textit{w/o Adapt} due to plan adaptation, both valid retrieval calls and sub-answer coverage improve substantially, demonstrating that \textbf{the dynamic adaptation mechanism effectively enables robust self-correction, allowing the model to recover from retrieval failures 
and identifying additional key evidence}.

\subsection{Effect of \textsc{SPlanRL} (RQ3)}
\label{subsec:splanrl_analysis}

\begin{table}[t]
\centering
\resizebox{0.85\linewidth}{!}{
\begin{tabular}{lcccc}
\toprule
\multirow{2}{*}{\textbf{Method}} & \multirow{2}{*}{\textbf{Metric}} & \multicolumn{3}{c}{\textbf{Hop Count}} \\
\cmidrule(lr){3-5}
 & & \textbf{2-hop} & \textbf{3-hop} & \textbf{4-hop} \\
\midrule
\multirow{2}{*}{\textsc{D$^2$Plan}-3B} & Pred. Hop & 2.03 & 2.70 & 3.20 \\
 & LasJ (\%) & 44.4 & 25.4 & 13.8 \\
\midrule
\multirow{2}{*}{w/o $R_p$} & Pred. Hop & 2.26 & 2.98 & 3.45 \\
 & LasJ (\%) & 39.7 & 24.1 & 12.8 \\
\bottomrule
\end{tabular}
}
\caption{Impact of Initial Planning Reward $R_p$ on question decomposition and final accuracy.}
\label{tab:rp_analysis}
\end{table}

\noindent \textbf{Initial Planning Reward Guides Concise and Focused Decomposition.} \quad
\label{para:rp_analysis}
To assess the impact of Initial Planning Reward $R_p$, we compare \textsc{D$^2$Plan} with \textit{w/o $R_p$} on subsets of MuSiQue with different hop counts (Table~\ref{tab:rp_analysis}).
Results indicate that \textit{w/o $R_p$} consistently predicts higher hop counts than \textsc{D$^2$Plan}, yet yields lower LasJ scores.
This suggests that removing $R_p$ induces over-decomposition, causing the model to generate extraneous sub-questions unrelated to the core problem.
For instance, given ``Where is Ulrich Walter's employer headquartered?'', \textit{w/o $R_p$} decomposes it into sub-questions including ``Who is Ulrich Walter?'', which, while tangentially related, does not directly contribute to answering the original question.
We sample 600 questions and employ \texttt{gemini-3-flash-preview} to detect such redundancies (see Appendix~\ref{app:prompts} for the prompt), followed by manual verification.
\textsc{D$^2$Plan} exhibits only 5 such cases, compared to 25 for \textit{w/o $R_p$}.
These findings confirm that $R_p$ guides the model toward concise decomposition strategies, thereby enhancing multi-hop reasoning performance.

\begin{table}[t]
\centering
\resizebox{0.73\linewidth}{!}{
\begin{tabular}{lcc}
\toprule
\textbf{Method} & \textbf{First VR (\%)} & \textbf{LasJ (\%)} \\
\midrule
\textsc{D$^2$Plan}-3B & 47.11 & 17.63 \\
\quad w/o $R_a$ & 44.79 & 14.49 \\
\bottomrule
\end{tabular}
}
\caption{Impact of Plan Adaptation Reward $R_a$ on MuSiQue. First VR: proportion of valid retrieval on the first query after revision
.}
\label{tab:ra_analysis}
\end{table}

\paragraph{Plan Adaptation Reward Steers Correct Plan Revisions.}
\label{para:ra_analysis}

To investigate the effect of Plan Adaptation Reward $R_a$, we compare \textsc{D$^2$Plan} with \textit{w/o $R_a$} on MuSiQue. 
We filter samples where plan revision occurred and report the valid retrieval rate (defined in \S\ref{subsec:retrieval_quality_analysis}) of the first generated query after the revision (First VR), along with the final LasJ accuracy (Table~\ref{tab:ra_analysis}).
Results show that \textsc{D$^2$Plan} significantly outperforms \textit{w/o $R_a$} in both First VR and LasJ.
This indicates that $R_a$ enables the model to effectively adapt the plan for self-correction when retrieval fails, leading to improved query reformulation and more accurate reasoning outcomes.

\subsection{Effect of Dual-Agent Architecture (RQ4)}
\label{subsec:dual_agent_analysis}

\begin{table}[t]
\centering
\resizebox{\linewidth}{!}{
\begin{tabular}{llccccc}
\toprule
\multirow{2}{*}{\textbf{Purifier}} & \multirow{2}{*}{\textbf{Metric}} & \multicolumn{5}{c}{\textbf{Hop Count}} \\
\cmidrule(lr){3-7}
 & & \textbf{2} & \textbf{3} & \textbf{4} & \textbf{5+} & \textbf{All} \\
\midrule
\multirow{3}{*}{w/o Purifier} & LasJ & 35.16 & 26.39 & 20.15 & 13.04 & 27.18 \\
 & SC & 4.22 & 5.00 & 5.57 & 5.77 & 4.89 \\
 & Len & 4451 & 5305 & 5921 & 6164 & 5180 \\
\midrule
\multirow{3}{*}{Qwen-7B-SFT} & LasJ & 36.77 & 27.78 & 22.39 & 20.65 & 29.49 \\
 & SC & 4.03 & 4.84 & 5.07 & 4.99 & 4.59 \\
 & Len & 1635 & 1976 & 2093 & 2124 & 1883 \\
\bottomrule
\end{tabular}
}
\caption{Impact of Purifier across different hop counts on FRAMES. LasJ: LLM-as-a-Judge (\%); SC: Average search calls; Len: Average context length (tokens).}
\label{tab:purifier_hop_analysis}
\end{table}

To validate the effectiveness of the dual-agent architecture, we train a \textit{Reasoner} baseline on trajectories synthesized from the same question set, but with raw retrieval results instead of Purifier-refined outputs.
This ensures the \textit{Reasoner} is not biased toward refined inputs, enabling a fair comparison.
We then evaluate the impact of adding the \textit{Purifier} (trained in \S\ref{subsec:sft}) on the FRAMES dataset.

\noindent \textbf{Greater Performance Gains on Harder Questions.} \quad
As shown in Table~\ref{tab:purifier_hop_analysis}, incorporating the \textit{Purifier} significantly improves performance across different hop counts while substantially reducing context length.
The advantage of the \textit{Purifier} is particularly pronounced on questions with 5 or more hops, where LasJ improves by 7.61\% and context length decreases by 65.5\%.
We attribute the substantially worse performance of \textit{w/o Purifier} to the accumulation of redundant information in raw retrieval results as question difficulty increases, which distracts the \textit{Reasoner} from focusing on critical information both within the current retrieval results and from prior reasoning steps.
The \textit{Purifier} effectively mitigates this issue by refining retrieval results and assessing their relevance.

\noindent \textbf{Dual-Agent Architecture Improves Inference Efficiency.} \quad
Although the \textit{Purifier} introduces additional inference steps, the end-to-end inference time is reduced.
Under the same computational budget, experiments show that using Qwen2.5-7B-SFT as the \textit{Purifier} reduces total inference time from 432s to 381s ($\sim$12\% reduction) on the FRAMES dataset. 
This efficiency gain is attributed to delegating the processing of lengthy retrieval results to the \textit{Purifier}, which substantially reduces the context length for the \textit{Reasoner}, thereby mitigating the quadratic computational cost of self-attention.

\subsection{Error Analysis (RQ5)}
\label{subsec:error_analysis}

\begin{figure}[t]
    \centering
    \includegraphics[width=0.85\linewidth]{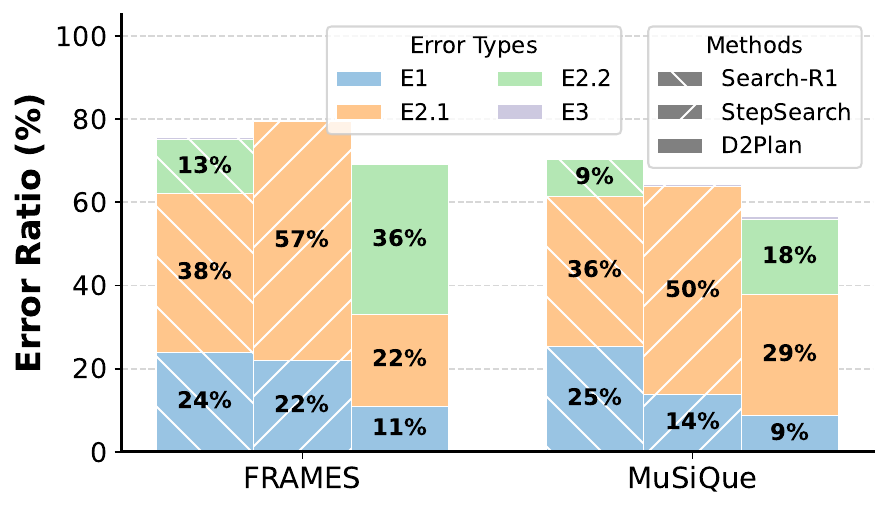}
    
    \caption{Error type distribution on FRAMES and MuSiQue. Percentages are calculated relative to the total number of questions, including correct predictions.}
    \label{fig:error_analysis_2}
\end{figure}
To understand how \textsc{D$^2$Plan} mitigates errors, we analyze error type distributions on FRAMES and MuSiQue, comparing it with Search-R1 and StepSearch. 
We divide \textbf{(E2) Reasoning Hijacking by Peripheral Evidence} into two subtypes:
(E2.1) the model reasons with irrelevant retrieval results without attempting remediation;
(E2.2) the model recognizes that the initial retrieval result is irrelevant and attempts query rewriting several times but still fails to obtain key evidence, resulting in an incorrect answer.
\footnote{E2.2 errors often stem from inherent retrieval system limitations, e.g., the corpus lacks relevant information entirely. See Appendix~\ref{app:e22_case} for a case study.}
Errors beyond E1 and E2 are categorized as \textbf{(E3) Other Errors}.
We sample 200 questions each from FRAMES and MuSiQue, using \texttt{gemini-3-pro-preview}~\cite{geminiteam2025geminifamilyhighlycapable} to identify the type of the first error in the reasoning trajectory.
We manually annotated 100 samples, achieving a Cohen's Kappa of 0.72, confirming Gemini's reliability.
As shown in Figure~\ref{fig:error_analysis_2}, \textsc{D$^2$Plan} achieves the lowest error rates by substantially reducing E1 and E2.1, validating the efficacy of dynamic global planning and robust resistance to peripheral evidence.
The relative rise in E2.2 suggests that as planning and peripheral evidence handling improve, failures shift toward inherent retrieval limitations and the constrained reasoning capabilities of the 7B model.

\section{Related Work}

\noindent \textbf{Planning in Retrieval-Augmented Generation.} \quad
Planning is crucial for RAG systems to handle complex multi-hop questions. 
Iterative RAG methods~\citep{shao2023enhancing,yao2023react,trivedi2022interleaving,liu2024ra,jiang2025retrieve,li2025search} interleave reasoning and retrieval but typically perform local planning, leading to intent drift in lengthy histories. 
PlanRAG~\citep{lee2024planrag} introduces a structured approach that generates plans and adjusts them via iterative re-planning. However, it relies on zero-shot prompting of models without parameter optimization. 

\noindent \textbf{Reinforcement Learning for Search-Augmented LLMs.} \quad
Recently, RL has become the dominant paradigm for training search-augmented LLMs~\cite{jin2025search,song2025r1,chen2025learning,sun2025zerosearchincentivizesearchcapability, shi2025search}, as it encourages autonomous exploration and better generalization than SFT~\cite{chu2025sft}.
Further research introduces step-wise process supervision~\citep{zheng-etal-2025-stepsearch,zhang-etal-2025-lets,luo2025globalrag}, reward density optimization~\citep{luo2025infoflow}, and self-correction mechanisms~\citep{wang2025erase} to enhance fine-grained reasoning and retrieval capabilities.
Despite these advances, current methods struggle with noisy and lengthy retrieval-augmented reasoning contexts.
Our method addresses these challenges by introducing a dual-agent dynamic global planning paradigm, optimized via a two-stage training framework.

\section{Conclusion}
We identify two prevalent failure modes of search-augmented LLMs in complex retrieval-augmented reasoning: Ineffective Search Chain Construction and Reasoning Hijacking by Peripheral Evidence. 
To address these issues, we introduce \textsc{D$^2$Plan}, a dual-agent dynamic planning paradigm where the Reasoner performs reasoning with explicit dynamic planning and the Purifier evaluates retrieval relevance and distills key information. 
A two-stage training framework comprising SFT cold-start and planning-oriented RL equips LLMs with the \textsc{D$^2$Plan} paradigm.
Extensive experiments demonstrate that \textsc{D$^2$Plan} enables more coherent query formulation and stronger resilience to peripheral evidence, achieving superior retrieval-augmented reasoning abilities.

\section{Limitations}
While \textsc{D$^2$Plan} demonstrates strong performance on open-domain QA benchmarks, several limitations remain.
First, although we achieve significant improvements on Qwen2.5-3B and Qwen2.5-7B models, due to computational cost constraints, we are unable to validate our method on larger language models. 
Second, due to cost limitations, we employ a local dense retriever (E5) with a static Wikipedia corpus rather than more powerful retrieval systems such as commercial search engines (e.g., Google). As revealed in our error analysis, a notable portion of failures (E2.2) may stem from retriever or corpus limitations. Integrating stronger retrieval infrastructure could potentially further enhance the performance of \textsc{D$^2$Plan}.
Third, our approach relies on a powerful teacher model to generate high-quality reasoning trajectories for supervised fine-tuning. This dependency may limit the applicability of our method in scenarios where access to such teacher models is restricted or costly.

\bibliography{custom}

\appendix
\section{\texorpdfstring{\textsc{D$^2$Plan}}{D2Plan} Workflow}
\label{app:workflow}

The inference workflow of \textsc{D$^2$Plan} is formalized in Algorithm~\ref{alg:d2plan_workflow}. Given an input question $Q$, the \textit{Reasoner} first assesses its complexity. For single-hop questions, it directly performs iterative retrieval and generates the answer. For multi-hop questions, it decomposes the question into an ordered sequence of sub-questions and solves them sequentially with dynamic plan adaptation.

\begin{algorithm*}[h!]
\caption{\textsc{D$^2$Plan} Inference Workflow}
\label{alg:d2plan_workflow}
\begin{algorithmic}[1]
\Input Question $Q$, Reasoner $\mathcal{R}$, Purifier $\mathcal{P}$, Retriever $\mathcal{E}$, Max Retrieval Attempts $K$, Max Revisions $M$
\Output Answer $a$
\State $type \leftarrow \mathcal{R}.\text{AssessComplexity}(Q)$ \Comment{Determine if single-hop or multi-hop}
\If{$type = \texttt{SingleHop}$}
    \State $status, info \leftarrow \text{IterativeRetrieval}(Q, \mathcal{R}, \mathcal{P}, \mathcal{E}, K, \emptyset)$ \Comment{Call Alg.~\ref{alg:iterative_retrieval}}
    \State $a \leftarrow \mathcal{R}.\text{GenerateAnswer}(Q, info, \emptyset)$
\Else
    \State $\mathcal{S} \leftarrow \mathcal{R}.\text{Decompose}(Q)$ \Comment{Generate initial sub-question sequence $[q_1, q_2, \ldots, q_n]$}
    \State $\mathcal{C} \leftarrow \emptyset$ \Comment{Initialize global context}
    \State $revision\_cnt \leftarrow 0$
    \State $i \leftarrow 1$
    \While{$i \le |\mathcal{S}|$}
        \State $q_i \leftarrow \mathcal{S}[i]$
        \State $status, info \leftarrow \text{IterativeRetrieval}(q_i, \mathcal{R}, \mathcal{P}, \mathcal{E}, K, \mathcal{C})$ \Comment{Call Alg.~\ref{alg:iterative_retrieval}}
        \If{$status = \texttt{Success}$}
            \State $a_i \leftarrow \mathcal{R}.\text{GenerateAnswer}(q_i, info, \mathcal{C})$
            \State $\mathcal{C} \leftarrow \mathcal{C} \cup \{(q_i, a_i, info)\}$
            \State $\mathcal{S}[i+1] \leftarrow \mathcal{R}.\text{RefineNextSubQuestion}(\mathcal{S}[i+1], \mathcal{C})$ \Comment{Plan Refinement}
            \State $i \leftarrow i + 1$
        \Else
            \If{$revision\_cnt < M$}
                \State $\mathcal{S}_{new} \leftarrow \mathcal{R}.\text{RevisePlan}(\mathcal{S}, i, \mathcal{C})$ \Comment{Plan Revision}
                \State $\mathcal{S} \leftarrow \mathcal{S}_{new}$
                \State $revision\_cnt \leftarrow revision\_cnt + 1$
            \Else
                \State $a_i \leftarrow \mathcal{R}.\text{GenerateAnswer}(q_i, \emptyset, \mathcal{C})$ \Comment{Answer with internal knowledge}
                \State $\mathcal{C} \leftarrow \mathcal{C} \cup \{(q_i, a_i, \emptyset)\}$
                \State $i \leftarrow i + 1$
            \EndIf
        \EndIf
    \EndWhile
    \State $a \leftarrow \mathcal{R}.\text{GenerateFinalAnswer}(Q, \mathcal{C})$
\EndIf
\State \Return $a$
\end{algorithmic}
\end{algorithm*}

\begin{algorithm*}[h!]
\caption{Iterative Retrieval with Purifier}
\label{alg:iterative_retrieval}
\begin{algorithmic}[1]
\Input Target $q$, Reasoner $\mathcal{R}$, Purifier $\mathcal{P}$, Retriever $\mathcal{E}$, Max Attempts $K$, Context $\mathcal{C}$
\Output Status, Purified Information
\State $\mathcal{H} \leftarrow \emptyset$ \Comment{Local retrieval history}
\For{$k = 1$ to $K$}
    \State $query_k \leftarrow \mathcal{R}.\text{GenerateQuery}(q, \mathcal{H}, \mathcal{C})$
    \State $\mathcal{D}_k \leftarrow \mathcal{E}.\text{Retrieve}(query_k)$ \Comment{Retrieve candidate documents}
    \State $info_k, relevance_k \leftarrow \mathcal{P}.\text{Purify}(q, \mathcal{D}_k)$ \Comment{Purifier judges relevance and extracts info}
    \State $\mathcal{H} \leftarrow \mathcal{H} \cup \{(query_k, info_k, relevance_k)\}$
    \If{$relevance_k = \texttt{Relevant}$}
        \State \Return \texttt{Success}, $info_k$
    \EndIf
\EndFor
\State \Return \texttt{Failure}, $\mathcal{H}$
\end{algorithmic}
\end{algorithm*}

\section{Trajectory Synthesis Strategy}
\label{app:trajectory_synthesis}

Given a question $Q$, the trajectory synthesis process is formalized in Algorithms \ref{alg:retrieval_loop}, \ref{alg:single_hop_synthesis}, and \ref{alg:multi_hop_synthesis}.

\paragraph{Iterative Retrieval Loop (Algorithm \ref{alg:retrieval_loop})}
This component serves as the fundamental building block for retrieving information. It executes a loop for up to $K$ attempts. In each iteration $k$:
\begin{enumerate}
    \item \textbf{Query Generation}: The teacher model $\mathcal{M}$ generates a reasoning thought $r_k$ and a search query $\mathbf{q}_k$ based on the current history $\mathcal{H}$. When the initial query fails to retrieve relevant information, we employ the query rewriting strategy (Table~\ref{tab:modify_query_prompt}) to refine the search query.
    \item \textbf{Retrieval and Purification}: The retriever fetches documents $\mathcal{D}_k$, and the \textit{Purifier} (acted by the teacher model, Table~\ref{tab:purifier_prompt}) extracts information $I_k$ with a relevance judgment $J_k$.
    \item \textbf{Termination}: If $J_k$ is \texttt{Relevant}, the loop terminates successfully returning the history $\mathcal{H}$. If the loop exhausts $K$ attempts without success, it returns \texttt{Failure}.
\end{enumerate}

\paragraph{Single-hop Question Processing (Algorithm \ref{alg:single_hop_synthesis})}
The process begins by assessing the question complexity. We denote the reasoning process for this assessment as $r_{type}$. For single-hop questions, it directly invokes the Iterative Retrieval Loop.
\begin{itemize}
    \item If the loop succeeds, the teacher model generates the final reasoning $r_{final}$ and answer $a$ based on the collected history (Table~\ref{tab:answer_subq_prompt}). The final trajectory includes the complexity assessment, retrieval history, and the final answer.
    \item If the loop fails, the synthesis process is terminated.
\end{itemize}

\paragraph{Multi-hop Question Processing (Algorithm \ref{alg:multi_hop_synthesis})}
The process begins by assessing the question complexity. We denote the reasoning process for this assessment as $r_{type}$. 
For multi-hop questions, the process follows the dynamic global planning paradigm:
\begin{enumerate}
    \item \textbf{Initialization}: The model decomposes $Q$ into an initial sub-question plan $\mathcal{P}$ (Table~\ref{tab:decompose_prompt}).
    \item \textbf{Execution Loop}: The model iterates through the sub-questions in $\mathcal{P}$. For each sub-question $q_i$:
    \begin{itemize}
        \item Before retrieval, we first judge whether external retrieval is necessary (Table~\ref{tab:judge_need_retrieval_prompt}). If so, it invokes the Iterative Retrieval Loop.
        \item \textbf{Success (Plan Refinement)}: If retrieval succeeds, the evidence is added to the global context $\mathcal{C}$. The model answers the sub-question (Table~\ref{tab:answer_subq_prompt}) and then refines the next sub-question based on $\mathcal{C}$ (Table~\ref{tab:update_subq_prompt}), making it more specific and self-contained.
        \item \textbf{Failure (Plan Revision)}: If retrieval fails, the model checks if the maximum revision count $M$ has been reached. If not, it triggers plan revision (Table~\ref{tab:replan_subquestions_prompt}) to revise the remaining sub-questions based on the current context and failure information. The execution resumes with the revised plan.
    \end{itemize}
    \item \textbf{Completion}: Once all sub-questions are solved, the model generates the final answer (Table~\ref{tab:conclude_prompt}).
\end{enumerate}

\begin{algorithm*}[h!]
\caption{Iterative Retrieval Loop}
\label{alg:retrieval_loop}
\begin{algorithmic}[1]
\Input Retrieval Target $q$, Teacher Model $\mathcal{M}$, Max Attempts $K$
\Output Status $\mathcal{S}$, Local Context $\mathcal{H}$
\State $\mathcal{H} \leftarrow \emptyset$
\For{$k = 1$ to $K$}
    \State $r_k, \mathbf{q}_k \leftarrow \mathcal{M}.\text{GenerateQuery}(q, \mathcal{H})$
    \State $\mathcal{D}_k \leftarrow \text{Retrieve}(\mathbf{q}_k)$
    \State $I_k, J_k \leftarrow \mathcal{M}.\text{Purify}(q, \mathcal{D}_k)$
    \State $\mathcal{H} \leftarrow \mathcal{H} \cup \{r_k, \mathbf{q}_k, I_k, J_k\}$
    \If{$J_k = \texttt{Relevant}$}
        \State \Return \texttt{Success}, $\mathcal{H}$
    \EndIf
\EndFor
\State \Return \texttt{Failure}, $\mathcal{H}$
\end{algorithmic}
\end{algorithm*}

\begin{algorithm*}[h!]
\caption{Trajectory Synthesis for Single-hop Questions}
\label{alg:single_hop_synthesis}
\begin{algorithmic}[1]
\Input Question $Q$, Teacher Model $\mathcal{M}$, Max Retrieval Attempts $K$
\Output Trajectory $\mathcal{T}$ or \texttt{Failure}
\State $r_{type} \leftarrow \mathcal{M}.\text{AssessComplexity}(Q)$ \Comment{Identify question complexity}
\State $\mathcal{S}, \mathcal{H} \leftarrow \text{RetrievalLoop}(Q, \mathcal{M}, K)$ \Comment{Call Alg. \ref{alg:retrieval_loop}}
\If{$\mathcal{S} = \texttt{Success}$}
    \State $r_{final}, a \leftarrow \mathcal{M}.\text{Answer}(Q, \mathcal{H})$ \Comment{Generate final answer}
    \State \Return $\mathcal{T} = \{Q, r_{type}\} \cup \mathcal{H} \cup \{r_{final}, a\}$
\Else
    \State \Return \texttt{Failure}
\EndIf
\end{algorithmic}
\end{algorithm*}

\begin{algorithm*}[h!]
\caption{Trajectory Synthesis for Multi-hop Questions}
\label{alg:multi_hop_synthesis}
\begin{algorithmic}[1]
\Input Question $Q$, Teacher Model $\mathcal{M}$, Max Revisions $M$, Max Retrieval Attempts $K$
\Output Trajectory $\mathcal{T}$ or \texttt{Failure}
\State $r_{type} \leftarrow \mathcal{M}.\text{AssessComplexity}(Q)$ \Comment{Identify question complexity}
\State $\mathcal{P} \leftarrow \mathcal{M}.\text{Decompose}(Q)$ \Comment{Generate initial sub-question sequence}
\State $\mathcal{C} \leftarrow \{Q, \mathcal{P}\}$ \Comment{Initialize global context}
\State $revision\_cnt \leftarrow 0$
\State $i \leftarrow 1$ \Comment{Index of current sub-question}
\While{$i \le |\mathcal{P}|$}
    \State $q_i \leftarrow \mathcal{P}[i]$
    \State $\mathcal{S}, \mathcal{H} \leftarrow \text{RetrievalLoop}(q_i, \mathcal{M}, K)$ \Comment{Call Alg. \ref{alg:retrieval_loop}}
    \If{$\mathcal{S} = \texttt{Success}$}
        \State $\mathcal{C} \leftarrow \mathcal{C} \cup \mathcal{H}$ \Comment{Accumulate evidence}
        \State $\mathcal{P} \leftarrow \mathcal{M}.\text{RefineNextSubQuestion}(\mathcal{P}, \mathcal{C})$ \Comment{Plan Refinement}
        \State $i \leftarrow i + 1$
    \Else
        \If{$revision\_cnt < M$}
            \State $\mathcal{P} \leftarrow \mathcal{M}.\text{RevisePlan}(Q, \mathcal{C})$ \Comment{Plan Revision}
            \State $revision\_cnt \leftarrow revision\_cnt + 1$
            \State \Comment{Resume from the first unsolved sub-question in new plan}
        \Else
            \State \Return \texttt{Failure}
        \EndIf
    \EndIf
\EndWhile
\State $a \leftarrow \mathcal{M}.\text{GenerateAnswer}(Q, \mathcal{C})$
\State $\mathcal{C} \leftarrow \mathcal{C} \cup \{a\}$
\State \Return $\mathcal{T} = \{r_{type}\} \cup \mathcal{C}$
\end{algorithmic}
\end{algorithm*}

\section{Training Details}
\label{app:training_details}

\paragraph{SFT Cold-Start}
We use \texttt{qwen-max-latest} \citep{qwen2025qwen25technicalreport} as the teacher model to synthesize reasoning trajectories from NQ, HotpotQA, MuSiQue (2-hop), and MuSiQue (3,4-hop) training sets, collecting 800, 1,600, 800, and 800 trajectories respectively. This yields 4,000 Reasoner samples and 11,255 Purifier samples. SFT is conducted using LLaMA-Factory~\citep{zheng2024llamafactory} with a learning rate of 1e-5, batch size of 64, for 3 epochs.
We set the maximum number of retrieval attempts to 3 and the maximum number of plan revisions to 1.

\paragraph{RL Training}
We sample 5,000, 10,000, 10,000, and 4,762 additional instances (non-overlapping with SFT data) from the NQ, HotpotQA, MuSiQue (2-hop), and MuSiQue (3,4-hop) training sets, totaling 29,762 samples. We adopt GRPO~\citep{shao2024deepseekmath} as the learning algorithm, implemented on verl~\citep{sheng2025hybridflow}. Training is conducted for 3 epochs with 5 rollouts per sample, using total/mini/micro batch sizes of 256/128/64, a learning rate of 1e-6, $\alpha = \beta = 0.1$, and discount factor $\lambda = 0.5$. 
We set the maximum number of retrieval attempts to 3.
Following common practice~\citep{jin2025empirical,jin2025search,song2025r1}, we mask out the tokens of search tool responses during training to prevent noise in the retrieval content from skewing the learning signal.

\section{Datasets}
\label{app:datasets}

We use publicly available QA benchmarks for training and evaluation. Table~\ref{tab:dataset_statistics} summarizes the statistics of all datasets used in our experiments.

\begin{table}[h]
\centering
\resizebox{0.85\linewidth}{!}{
\begin{tabular}{lcc}
\toprule
\textbf{Dataset} & \textbf{Train} & \textbf{Dev/Test} \\
\midrule
NQ & 79,168 & 3,610 \\
SimpleQA & - & 4,326 \\
HotpotQA & 90,447 & 7,405 \\
2WikiMultihopQA & 15,000 & 12,576 \\
MuSiQue & 19,938 & 2,417 \\
FRAMES & - & 824 \\
\bottomrule
\end{tabular}
}
\caption{Statistics of datasets used in our experiments. ``-'' indicates the split is not used or not available.}
\label{tab:dataset_statistics}
\end{table}

\paragraph{Data Sources}
All datasets except SimpleQA and FRAMES are obtained from FlashRAG~\citep{FlashRAG}, a modular toolkit for retrieval-augmented generation research that provides standardized dataset preprocessing and evaluation pipelines.
These datasets are publicly available and do not contain personally identifiable information or offensive content.

\section{Ablation Study Implementation Details}
\label{app:ablation_details}

For the removal of the Dynamic Global Planning Paradigm (\textit{w/o Plan\&Adapt}), we do not use the reasoning trajectory synthesis method described in \S\ref{subsec:sft}. 
Instead, while retaining the dual-agent architecture, we employ the prompt provided in Table~\ref{tab:prompt_wo_global_planning} to guide the teacher model in rolling out solutions. 
We then apply rejection sampling to retain only those reasoning trajectories that yield correct answers, which are subsequently used as SFT data. 
In the RL phase, we optimize using only format and answer rewards, omitting all planning-related reward designs.

For \textit{w/o Adapt}, we disable the plan adaptation mechanism, meaning the model generates an initial plan but lacks the ability to replan or perform iterative retrieval when retrieval results are irrelevant. 
During the RL phase, we use the format reward, answer reward, and the Initial Planning Reward $R_p$.

We conduct SFT training using reasoning trajectories synthesized from the same question set.

For \textit{w/o $R_p$} and \textit{w/o $R_a$}, we remove the initial planning reward and adaptation reward respectively, while keeping all other components unchanged.

\section{Format Reward Constraints}
\label{app:format_constraints}

The format reward $R_f$ enforces the following constraints to ensure parsable reasoning paths:

\begin{enumerate}
    \item \textbf{Answer Tag Constraint}: The model output must contain exactly one pair of \texttt{<answer>} and \texttt{</answer>} tags, enclosing the final answer.
    
    \item \textbf{Plan Tag Constraint}: For multi-hop questions, the output must contain \texttt{<plan>} and \texttt{</plan>} tags that enclose the initial decomposition of sub-questions.
    
    \item \textbf{Sub-question Identifier Constraint}: Within the plan, each sub-question must be marked with identifiers "\texttt{\#Q\_i: }" (where $i$ is the sub-question index), enabling systematic tracking of the solving process.

    \item \textbf{Refinement Tag Constraint}: When refining a sub-question after solving its dependencies, the model must use \texttt{<Updated\_\#Q\_i>} and \texttt{</Updated\_\#Q\_i>} tags to mark the refined version.
    
    \item \textbf{Revision Tag Constraint}: When triggering plan revision after retrieval failures, the model must use \texttt{<Replan>} and \texttt{</Replan>} tags to enclose the revised plan.
\end{enumerate}

These constraints ensure that the reasoning trajectory can be automatically parsed via regular expressions to compute the Initial Planning Reward $R_p$ (by extracting the initial plan and counting sub-questions) and the Plan Adaptation Reward $R_a$ (by detecting refinement tags, revision tags, and iterative retrieval patterns).

\section{Comparison of Different Purifier Agents}
\label{app:purifier_comparison}

This section extends the analysis in \S\ref{sec:results_analysis} by comparing different Purifier configurations. We use the same experimental setup described in the main text.

\begin{table}[t]
\centering
\resizebox{0.9\linewidth}{!}{
\begin{tabular}{lcccc}
\toprule
\textbf{Purifier} & LasJ & SC & Len & Time \\
\midrule
Search-R1 & 19.78 & 3.01 & 2883 & - \\
\midrule
w/o Purifier & 27.18 & 4.89 & 5180 & 432 \\
\midrule
Qwen2.5-7B-Instruct & 26.70 & 5.06 & 1724 & 359\\
Qwen2.5-32B-Instruct & 28.88 & 4.89 & 1758 & 653\\
Qwen-Max & 32.16 & 4.65 & 2073 & - \\
\midrule
Qwen2.5-3B-SFT & 27.30 & 4.70 & 1979 & 345\\
Qwen2.5-7B-SFT & 29.49 & 4.59 & 1883 & 381\\
\bottomrule
\end{tabular}
}
\caption{Performance comparison of different Purifier Agents on FRAMES. LasJ: LLM-as-a-Judge (\%); SC: Number of search calls; Len: Context length (tokens); Time: Total inference time (seconds). Search-R1 serves as a baseline without the dual-agent architecture.}
\label{tab:purifier_performance}
\end{table}

As shown in Table~\ref{tab:purifier_performance}, we compare three categories of Purifier configurations: (1) no Purifier baseline, (2) general-purpose instruction-tuned models (Qwen2.5-7B/32B-Instruct and Qwen-Max), and (3) our task-specific fine-tuned models (Qwen2.5-3B/7B-SFT). Search-R1 serves as an external baseline representing the single-agent paradigm.

\paragraph{Task-Specific Fine-Tuning Enables Smaller Models to Serve as Effective Purifiers.}
Among general-purpose models, larger Purifiers generally yield better performance, with Qwen-Max achieving the highest LasJ score of 32.16\%.
However, the general-purpose Qwen2.5-7B-Instruct fails to effectively perform the Purifier task, achieving only 26.70\% LasJ---even lower than the no-Purifier baseline (27.18\%). This indicates that smaller general-purpose models lack the capability to accurately judge relevance and extract key information from noisy retrieval results.
In contrast, our fine-tuned Qwen2.5-7B-SFT achieves a LasJ score of 29.49\%, which not only recovers from this deficiency but also surpasses Qwen2.5-32B-Instruct (28.88\%) despite having only 22\% of its parameters.
This demonstrates that task-specific fine-tuning is essential for enabling moderate-sized models to serve as effective Purifiers, offering a practical trade-off between performance and computational cost.

\paragraph{Inference Efficiency Details.}
We deploy the \textit{Purifier} using vLLM's server mode on an 8$\times$A800 node. Deploying an additional 7B \textit{Purifier} incurs approximately 20\% additional GPU memory overhead.
The general-purpose Qwen2.5-32B-Instruct not only achieves lower accuracy (28.88\% vs. 29.49\%) but also incurs significantly higher inference time (653s) due to its larger model size.
This comparison highlights the value of task-specific fine-tuning: our Qwen2.5-7B-SFT, with only 22\% of the parameters, outperforms the 32B general-purpose model in both accuracy and efficiency.

\section{Case Study}
\label{appendix:case}

\newcommand{\think}[1]{\textcolor{blue}{\texttt{<think>}} #1 \textcolor{blue}{\texttt{</think>}}}
\newcommand{\search}[1]{\textcolor{cyan}{\texttt{<search>}} #1 \textcolor{cyan}{\texttt{</search>}}}
\newcommand{\information}[1]{\textcolor{brown}{\texttt{<information>}} #1 \textcolor{brown}{\texttt{</information>}}}
\newcommand{\answer}[1]{\textcolor{purple}{\texttt{<answer>}} #1 \textcolor{purple}{\texttt{</answer>}}}

\newcommand{\planopen}{\textcolor{teal}{\texttt{<plan>}}}
\newcommand{\planclose}{\textcolor{teal}{\texttt{</plan>}}}
\newcommand{\toolcallopen}{\textcolor{orange}{\texttt{<tool\_call>}}}
\newcommand{\toolcallclose}{\textcolor{orange}{\texttt{</tool\_call>}}}
\newcommand{\toolresponseopen}{\textcolor{olive}{\texttt{<tool\_response>}}}
\newcommand{\toolresponseclose}{\textcolor{olive}{\texttt{</tool\_response>}}}
\newcommand{\finalansweropen}{\textcolor{violet}{\texttt{<answer>}}}
\newcommand{\finalanswerclose}{\textcolor{violet}{\texttt{</answer>}}}
\newcommand{\subq}[1]{\textcolor{blue}{\texttt{\#Q\_#1}}}
\newcommand{\suba}[1]{\textcolor{magenta}{\texttt{\#A\_#1}}}

In this section, we present comparative case studies derived from the interaction trajectories of \textbf{Search-R1} and our proposed \textbf{\textsc{D$^2$Plan}}. Specifically, Tables~\ref{tab:case_searchr1} and \ref{tab:case_d2plan} illustrate a scenario where \textbf{Search-R1} exhibits E1 (Deficient Search Chain) by generating a flawed query sequence that diverges from the original reasoning objective---instead of querying the discoverer's birthdate, it directly searches for a holiday, leading to an incorrect answer. In contrast, our proposed \textbf{\textsc{D$^2$Plan}} successfully solves the problem through explicit global planning and dynamic sub-question refinement, maintaining alignment with the original reasoning objective throughout the process.

\begin{table*}[htbp]
\centering
\footnotesize
\begin{tabular}{p{15cm}}
\toprule
\textbf{Question}: The Basibasy mine is located in Madagascar. This mine is abundant in a specific chemical element that was discovered for the first time in 1791. The person who discovered this element was born on what is now known as a major US holiday - what holiday is this? \\
\textbf{Golden Answer}: Christmas \\
\midrule
\textbf{Search-R1} \textcolor{red}{(Incorrect)}: \\
\search{Basibasy mine Madagascar} \\
\information{Doc 1(Title: ``Ambatovy mine'') Ambatovy, located in Madagascar, off the east coast of Africa, is one of the world's largest lateritic nickel mines with a production capacity of 60,000 tonnes of nickel... Doc 2(Title: Maniry) Maniry is a town and commune in southwestern Madagascar. It belongs to the district of Ampanihy... Doc 3(Title: ``Economy of Madagascar'') coal at Sakoa and nickel near Tamatave. The Ambatovy nickel mine is a huge operation... Doc 4(Title: ``Ambatovy mine'') Bank CSRM report estimate the impact of the mining industry as two to five indirect jobs... Doc 5(Title: ``Mining industry of Madagascar'') of mining Code. Minerals, both industrial and metallic, are numerous including ilmenite, graphite...} \\
Based on the information provided, the Basibasy mine seems to be referred to as Ambatovy mine, which is located in Madagascar and is a large nickel mine. Now, let's find the element discovered in 1791. \\
\search{element discovered in 1791} \\
\information{Doc 1(Title: ``Group 9 element'') regia. Iridium was discovered in a very similar way in 1804 by Smithson Tennant... Doc 2(Title: Titanium) Titanium was discovered in 1791 by the clergyman and amateur geologist, William Gregor, as an inclusion of a mineral in Cornwall, Great Britain... Doc 3(Title: ``Group 3 element'') the current heaviest element known, oganesson. In 1787, Swedish part-time chemist Carl Axel Arrhenius... Doc 4(Title: ``Anders Gustaf Ekeberg'') Anders Gustaf Ekeberg was a Swedish chemist who discovered tantalum in 1802... Doc 5(Title: ``Group 9 element'') Meitnerium has not been isolated in pure form...} \\
From the information provided, it seems William Gregor discovered the element titanium in 1791. Now, let's find the US holiday related to the birth date provided. \\
\search{person born on Thanksgiving} \\
\information{Doc 1(Title: ``Thanksgiving (United States)'') Thanksgiving is a national holiday in the United States, celebrated on the fourth Thursday of November. It originated as a harvest festival... Doc 2(Title: Thanksgiving) Thanksgiving Day is a national holiday celebrated on various dates in Canada, the United States... Doc 3(Title: Thanksgiving) after 1700, late autumn Thanksgiving celebrations became commonplace... Doc 4(Title: Thanksgiving) beginning of the 19th century, coinciding with and eventually superseding the existing holiday... Doc 5(Title: ``Preterm birth'') Ahmad resuscitated the baby after she was born...} \\
The information states that Thanksgiving is a national holiday in the United States, celebrated on the fourth Thursday of November. The person who discovered titanium, William Gregor, was born on a day that is now a US holiday. Therefore, the holiday in question is likely Thanksgiving. \\
\answer{Thanksgiving} \\
\bottomrule
\end{tabular}
\caption{Case study: Search-R1 exhibits E1 (Deficient Search Chain) by generating a flawed query sequence. Instead of querying William Gregor's birthdate, it directly searches for ``person born on Thanksgiving'', diverging from the correct reasoning objective and leading to an incorrect answer.}
\label{tab:case_searchr1}
\end{table*}

\begin{table*}[htbp]
\centering
\footnotesize
\begin{tabular}{p{15cm}}
\toprule
\textbf{Question}: The Basibasy mine is located in Madagascar. This mine is abundant in a specific chemical element that was discovered for the first time in 1791. The person who discovered this element was born on what is now known as a major US holiday - what holiday is this? \\
\textbf{Golden Answer}: Christmas \\
\midrule
\textbf{\textsc{D$^2$Plan}} \textcolor{green}{(Correct)}: \\
This is a multi-hop question involving multiple steps: identifying the chemical element discovered in 1791 at the Basibasy mine in Madagascar, determining the discoverer's birth date, and linking it to a significant U.S. holiday. Each step builds on the previous one, requiring sequential reasoning. \\
\planopen \\
\subq{1}: What chemical element was discovered for the first time in 1791 and is abundant in the Basibasy mine in Madagascar? \\
\subq{2}: On what day was the person who discovered the element identified in \suba{1} born? \\
\subq{3}: What major US holiday is associated with the birthdate identified in \suba{2}? \\
\planclose \\
To proceed, I'll first identify the chemical element linked to both the Basibasy mine and its discovery in 1791. External verification will confirm whether this aligns with historical records of discoveries and elemental abundances. \\
\toolcallopen \\
\{``name'': ``search'', ``arguments'': \{``query'': ``chemical element discovered in 1791 in Madagascar'', ``question'': ``What chemical element was discovered for the first time in 1791 and is abundant in the Basibasy mine in Madagascar?''\}\} \\
\toolcallclose \\
\toolresponseopen \\
\{``result'': ``Doc 1: [TARGET\_INFO\_EXTRACTED] Titanium was discovered in 1791 by the clergyman and amateur geologist William Gregor; Titanium is abundant in the Basibasy mine in Madagascar''\} \\
\toolresponseclose \\
The document states that titanium was discovered in 1791 and is abundant in the Basibasy mine in Madagascar, directly answering the question. \\
\suba{1}: Titanium \\
I'll now determine the discoverer's birthdate: \\
<updated\_\subq{2}> \\
On what day was William Gregor, the person who discovered titanium, born? \\
</updated\_\subq{2}> \\
The prior answer identifies titanium but does not include details about William Gregor's birthdate. External information is needed to address this sub-question. \\
\toolcallopen \\
\{``name'': ``search'', ``arguments'': \{``query'': ``William Gregor birth date'', ``question'': ``On what day was William Gregor, the person who discovered titanium, born?''\}\} \\
\toolcallclose \\
\toolresponseopen \\
\{``result'': ``Doc 1: [TARGET\_INFO\_EXTRACTED] William Gregor was born on 25 December 1761''\} \\
\toolresponseclose \\
The document provides William Gregor's birth date, including the day. \\
\suba{2}: 25 December \\
I'll now determine which major U.S. holiday corresponds to this date. \\
<updated\_\subq{3}>\\
What major US holiday is celebrated on December 25? \\
</updated\_\subq{3}>\\
The answer is common knowledge: Christmas Day, a widely recognized national holiday in the United States. \\
\suba{3}: Christmas Day \\
\finalansweropen \\
Christmas Day \\
\finalanswerclose \\
\bottomrule
\end{tabular}
\caption{Case study: \textsc{D$^2$Plan} succeeds through structured planning and dynamic sub-question refinement, effectively mitigating E1 (Deficient Search Chain). By explicitly decomposing the problem and dynamically updating \subq{2} with the discovered entity (William Gregor), it avoids diverging from the original reasoning objective and correctly queries the birthdate, arriving at the correct answer.}
\label{tab:case_d2plan}
\end{table*}

Additionally, Tables~\ref{tab:case_searchr1_replan} and \ref{tab:case_replan} demonstrate how \textbf{\textsc{D$^2$Plan}}'s plan revision mechanism corrects an initially flawed reasoning direction. In this example, \textbf{Search-R1} incorrectly identifies ``line'' as a synonym for ``leash'' and then erroneously concludes that ``Lithium'' sounds like ``line'', leading to an incorrect answer. In contrast, \textbf{\textsc{D$^2$Plan}}'s initial decomposition leads to semantic drift---retrieving engineering ``dog'' tools instead of pet walking equipment. After two failed retrieval attempts, the model triggers plan revision that merges the first two sub-questions, directly querying for synonyms of dog walking tools. This successfully redirects the reasoning and arrives at the correct answer ``Lead'' (both a synonym for leash and a chemical element).

\begin{table*}[htbp]
\centering
\footnotesize
\begin{tabular}{p{15cm}}
\toprule
\textbf{Question}: Which element of the periodic table is a homonym of a synonym of a tool commonly used in dog walking? \\
\textbf{Golden Answer}: Lead \\
\midrule
\textbf{Search-R1} \textcolor{red}{(Incorrect)}: \\
To solve this problem, I'll need to break it down step by step: 1. Identify a tool commonly used in dog walking. 2. Find a synonym for that tool. 3. Find an element in the periodic table that is a homonym of that synonym. \\
Step 1: A tool commonly used in dog walking is a leash or a lead. \\
Step 2: A synonym for ``leash'' or ``lead'' could be ``line.'' \\
Step 3: I need to find an element in the periodic table that is a homonym of ``line.'' \\
\search{homonym of line periodic table} \\
\information{Doc 1(Title: ``Homology (chemistry)'') Homology (chemistry) In chemistry, homology is the appearance of homologues. A homologue is a compound belonging to a series of compounds differing from each other by a repeating unit... Doc 2(Title: ``Periodic table'') The periodic table is a tabular arrangement of the chemical elements, ordered by their atomic number...} \\
From the information provided, I can see that there is no direct mention of an element that is a homonym of ``line.'' However, the periodic table contains elements with various names. I will search for elements with names that sound like ``line.'' \\
\search{elements periodic table names sound like line} \\
\information{Doc 1(Title: ``Periodic table'') The periodic table is a tabular arrangement of the chemical elements... Doc 2(Title: ``Naming of chemical elements'') Chemical elements may be named from various sources: sometimes based on the person who discovered it, or the place it was discovered...} \\
There is no element in the periodic table that is a homonym of ``line.'' I will try searching for elements with names that sound like ``line.'' \\
\search{elements periodic table names sound like line or lean} \\
\information{Doc 1(Title: ``Periodic table'') The periodic table is a tabular arrangement of the chemical elements... Doc 2(Title: ``Chemical element'') The terms ``light'' and ``heavy'' are sometimes also used informally to indicate relative atomic numbers...} \\
After searching, I found that there is an element in the periodic table named ``Lithium,'' which sounds like ``line.'' Therefore, the answer is: \\
\answer{Lithium} \\
\bottomrule
\end{tabular}
\caption{Case study: Search-R1 incorrectly identifies ``line'' as a synonym for ``leash'' and then erroneously concludes that ``Lithium'' sounds like ``line.'' The model fails to recognize that ``lead'' is both a synonym for leash and a chemical element, leading to an incorrect answer.}
\label{tab:case_searchr1_replan}
\end{table*}

\begin{table*}[htbp]
\centering
\footnotesize
\begin{tabular}{p{15cm}}
\toprule
\textbf{Question}: Which element of the periodic table is a homonym of a synonym of a tool commonly used in dog walking? \\
\textbf{Golden Answer}: Lead \\
\midrule
\textbf{\textsc{D$^2$Plan}} \textcolor{green}{(Correct)}: \\
This is a multi-hop question involving two steps: identifying a tool used in dog walking and finding an element whose name matches a synonym of that tool. The phrasing introduces ambiguity and requires sequential reasoning. \\
\planopen \\
\subq{1}: What is a tool commonly used in dog walking? \\
\subq{2}: What is a synonym of the tool identified in \suba{1}? \\
\subq{3}: Which element of the periodic table is a homonym of the synonym identified in \suba{2}? \\
\planclose \\
First, I'll identify a tool frequently used in dog walking. External knowledge is needed since no prior context provides this information. \\
\toolcallopen \\
\{``name'': ``search'', ``arguments'': \{``query'': ``common tool used in dog walking'', ``question'': ``What is a tool commonly used in dog walking?''\}\} \\
\toolcallclose \\
\toolresponseopen \\
\{``result'': ``[NO\_TARGET\_INFO\_FOUND]\textbackslash nSummary of retrieved documents:\textbackslash n[Doc 1]: Dog walking involves a person walking with a dog... [Doc 2]: Dog walking refers to the act of walking a dog... [Doc 3]: In engineering, a dog is a tool or part... [Doc 4]: Engineering tools called `dogs' include feed dogs... [Doc 5]: A bench dog is a woodworking accessory...''\} \\
\toolresponseclose \\
The question asks about a tool used in dog walking. However, the documents describe `dog' as an engineering term referring to tools like dog clutches, feed dogs in sewing machines, and bench dogs in woodworking---not as a pet-related term. The documents lack sufficient semantic alignment to answer the question. \\
Let me modify my search query using Entity Enrichment / Correction strategy. \\
\toolcallopen \\
\{``name'': ``search'', ``arguments'': \{``query'': ``engineering tool named dog --- dog clutch, feed dog in sewing machine, or bench dog in woodworking'', ``question'': ``What is a tool commonly used in dog walking?''\}\} \\
\toolcallclose \\
\toolresponseopen \\
\{``result'': ``[NO\_TARGET\_INFO\_FOUND]\textbackslash nSummary of retrieved documents:\textbackslash n[Doc 1]: In engineering, a dog is a tool or part... [Doc 2]: A dog in engineering refers to a tool or mechanism... [Doc 3]: A lathe dog, also known as a lathe carrier... [Doc 4]: A bench dog is a woodworking accessory... [Doc 5]: Shutter dogs are wrought-iron or cast-iron devices...''\} \\
\toolresponseclose \\
The documents still describe `dog' in engineering contexts but do not mention dog walking or related activities. (No relevant docs found after 2 attempts.) \\
\textcolor{red}{\textbf{Plan revision triggered due to repeated irrelevant retrieval.}} \\
\texttt{<replan>} \\
\subq{1}: What is a synonym of a tool commonly used in dog walking? \\
\subq{2}: Which element of the periodic table is a homonym of the synonym identified in \suba{1}? \\
\texttt{</replan>} \\
The first step is addressing \subq{1}. External information is required since the context does not specify a tool for dog walking or its synonyms. \\
\toolcallopen \\
\{``name'': ``search'', ``arguments'': \{``query'': ``What is a synonym for leash or harness, a tool commonly used in dog walking?'', ``question'': ``What is a synonym of a tool commonly used in dog walking?''\}\} \\
\toolcallclose \\
\toolresponseopen \\
\{``result'': ``Doc 1: [TARGET\_INFO\_EXTRACTED] A leash is also called a lead, lead line or tether''\} \\
\toolresponseclose \\
The document states that a leash is also called a lead, lead line, or tether---synonyms for a tool used in dog walking. \\
\suba{1}: Lead \\
I'll now determine which chemical element shares its name with this synonym. \\
<updated\_\subq{2}> \\
Which element of the periodic table is a homonym of the synonym ``lead''? \\
</updated\_\subq{2}> \\
The synonym identified was `lead', which is already a chemical element (Pb). No external retrieval is necessary. \\
\suba{2}: Lead \\
\finalansweropen \\
Lead \\
\finalanswerclose \\
\bottomrule
\end{tabular}
\caption{Case study: \textsc{D$^2$Plan}'s plan revision mechanism corrects semantic drift through dynamic replanning.}
\label{tab:case_replan}
\end{table*}

\section{Error Analysis: E2.1 vs. E2.2 Error Case}
\label{app:e22_case}
Tables~\ref{tab:case_e22_searchr1} and \ref{tab:case_e22} illustrate the difference between E2.1 and E2.2 errors, comparing the behaviors of Search-R1 and \textsc{D$^2$Plan} on the same question. 

As shown in Table~\ref{tab:case_e22_searchr1}, Search-R1 successfully retrieves Hod Lisenbee's place of death (Clarksville) but fails to find the distance information in the second retrieval. Critically, \textbf{Search-R1 does not attempt to reformulate the query or re-retrieve}---instead, it directly hallucinates an answer (``approximately 60 miles'') based on its internal knowledge, which is incorrect. This exemplifies \textbf{E2.1 errors}: the model reasons with irrelevant retrieval results without attempting any remediation.

In contrast, as shown in Table~\ref{tab:case_e22}, \textsc{D$^2$Plan} demonstrates more robust behavior: when the initial retrieval for the distance fails, the model appropriately recognizes the irrelevance and attempts query reformulation with enriched entities (adding constraints like ``miles'', ``km'', and authoritative sources). However, despite multiple retrieval attempts with refined queries, the retrieval system cannot locate the required distance information, causing the model to hallucinate an incorrect answer (95 miles instead of 45 miles). This exemplifies \textbf{E2.2 errors}: the model attempts remediation but fails due to inherent corpus limitations. 

Upon manual inspection, we traced the ground-truth answer to the Wikipedia article for Clarksville, Tennessee (\url{https://en.wikipedia.org/wiki/Clarksville,_Tennessee}), which states: ``Clarksville is located on the northwest edge of the Highland Rim, which surrounds the Nashville Basin, and is 45 miles (72 km) northwest of Nashville.''
However, the corresponding document in our corpus does not contain the critical distance information ``45 miles (72 km)'':

\begin{quote}
\small
\texttt{\{``id'': ``14985934'', ``contents'': ``"Clarksville, Tennessee" was rededicated. In addition to restoring the 1879 courthouse and plazas, the county built a new courts center on the north side for the court operations. According to the United States Census Bureau, the city has a total area of , of which is land and (0.71\%) is covered by water. \textbf{Clarksville is located on the northwest edge of the Highland Rim, which surrounds the Nashville Basin, and is northwest of Nashville.} Fort Campbell North is a census-designated place (CDP) in Christian County, Kentucky. It contains most of the housing for the Fort Campbell Army base. The population was 14,338''\}}
\end{quote}

\noindent As shown, the corpus document retains only the directional phrase ``northwest of Nashville'' without the numeric distance, making it impossible for the model to retrieve the correct answer.
This comparison highlights the key difference between E2.1 and E2.2 errors: Search-R1 exhibits E2.1 behavior by immediately resorting to hallucination without any remediation effort, whereas \textsc{D$^2$Plan} demonstrates superior self-correction behavior by actively attempting to reformulate queries---though it ultimately fails due to corpus limitations (E2.2). This case study illustrates how \textsc{D$^2$Plan}'s dynamic planning paradigm shifts errors from model capability deficiencies (E2.1) to inherent retrieval system limitations (E2.2).

\begin{table*}[htbp]
\centering
\footnotesize
\begin{tabular}{p{15cm}}
\toprule
\textbf{Question}: How far is Hod Lisenbee's place of death in TN from Nashville? \\
\textbf{Golden Answer}: 45 \\
\midrule
\textbf{Search-R1} \textcolor{red}{(Incorrect --- E2.1 Error)}: \\
\think{I need to find out the place of death for Hod Lisenbee first.} \\
\search{place of death for Hod Lisenbee} \\
\information{Doc 1(Title: ``Hod Lisenbee'') He died at age 89 on November 14, 1987, in Clarksville and is buried at the Liberty Presbyterian Church Cemetery. In 1998, the Clarksville City Council renamed a portion of Dover Road (Highway 79) near the Dover Crossing interception as Hod Lisenbee memorial highway... Doc 2(Title: ``Hod Lisenbee'') retired from baseball. Hod Lisenbee was manager and half-owner of the Clarksville Colts club from 1946 to 1948... Lisenbee lived in his hometown of Clarksville from the fall of 1945, until his death in 1987... Doc 3(Title: ``Hod Lisenbee'') from the Cincinnati Reds and pitched 31 games for the Reds, as a reliever... Following his release from the Reds, he returned to his hometown of Clarksville...} \\
From the information, Hod Lisenbee's place of death is in Clarksville. Now I need to find out how far Clarksville is from Nashville. \\
\search{distance from Clarksville to Nashville} \\
\information{Doc 1(Title: ``Clarksville, Tennessee'') Clarksville, Tennessee Clarksville is the county seat of Montgomery County, Tennessee, United States. It is the fifth-largest city in the state behind Nashville, Memphis, Knoxville, and Chattanooga. The city had a population of 132,929 at the 2010 census... Doc 2(Title: ``Nashville Metropolitan Transit Authority'') public hearings on April 24, 2012 regarding proposed regional express bus service from Clarksville to Nashville... Doc 3(Title: ``Clarksville, Tennessee'') Pleasant Academy. By 1819, the newly established town had 22 stores... Trade via land also grew as four main dirt roads were established, two to Nashville...} \\
\textcolor{orange}{\textbf{From the information, Clarksville is the fifth-largest city in Tennessee, and it is located in Montgomery County. Nashville is the capital of Tennessee and the county seat of Davidson County. The distance from Clarksville to Nashville is approximately 60 miles.}} \\
\answer{60 miles} \\
\bottomrule
\end{tabular}
\caption{Case study illustrating \textbf{E2.1 error} of Search-R1: the model fails to retrieve the distance information and \textbf{does not attempt to reformulate the query}. Instead, it directly hallucinates an incorrect answer (60 miles) based on internal knowledge, demonstrating lack of self-correction capability.}
\label{tab:case_e22_searchr1}
\end{table*}

\begin{table*}[htbp]
\centering
\footnotesize
\begin{tabular}{p{15cm}}
\toprule
\textbf{Question}: How far is Hod Lisenbee's place of death in TN from Nashville? \\
\textbf{Golden Answer}: 45 \\
\midrule
\textbf{\textsc{D$^2$Plan}} \textcolor{red}{(Incorrect --- E2.2 Error)}: \\
This is a multi-hop question requiring two steps: identifying Hod Lisenbee's place of death in Tennessee and calculating its distance from Nashville. Each step depends on resolving the prior one. \\
\planopen \\
\subq{1}: Where did Hod Lisenbee die? \\
\subq{2}: How far is \suba{1} (Hod Lisenbee's place of death) from Nashville, TN? \\
\planclose \\
I'll first determine where Hod Lisenbee died. This information isn't provided in the question or prior context, so it must be retrieved externally. \\
\toolcallopen \\
\{``name'': ``search'', ``arguments'': \{``query'': ``Where did Hod Lisenbee die?'', ``question'': ``Where did Hod Lisenbee die?''\}\} \\
\toolcallclose \\
\toolresponseopen \\
\{``result'': ``Doc 1: [TARGET\_INFO\_EXTRACTED] Hod Lisenbee died in Clarksville''\} \\
\toolresponseclose \\
The document states that Hod Lisenbee died in Clarksville. \\
\suba{1}: Clarksville \\
With Clarksville identified as the location of Hod Lisenbee's death, I'll now calculate the distance from there to Nashville, TN. \\
<updated\_\subq{2}> \\
How far is Clarksville, TN from Nashville, TN? \\
</updated\_\subq{2}> \\
This requires determining the geographical distance between the two cities, which hasn't been covered yet. \\
\toolcallopen \\
\{``name'': ``search'', ``arguments'': \{``query'': ``distance from Clarksville TN to Nashville TN'', ``question'': ``How far is Clarksville, TN from Nashville, TN?''\}\} \\
\toolcallclose \\
\toolresponseopen \\
\{``result'': ``[NO\_TARGET\_INFO\_FOUND] The Clarksville, TN--KY metropolitan statistical area consists of Montgomery and Stewart Counties in Tennessee, and Christian and Trigg Counties in Kentucky; The city distance between Clarksville, TN from Nashville is not provided...''\} \\
\toolresponseclose \\
The document mentions the Clarksville, TN--KY metropolitan statistical area but does not provide the actual distance. Thus, it lacks sufficient information to answer the question directly. \\
\textcolor{orange}{\textbf{The retrieved documents are not relevant. I'll modify my search query.}} The previous query returned contextual details about the Clarksville-Nashville region but failed to yield the numeric distance due to missing explicit entity constraints like units (``miles'', ``km'') and authoritative source signals. \\
\toolcallopen \\
\{``name'': ``search'', ``arguments'': \{``query'': ``What is the straight-line driving distance in miles between Clarksville, Tennessee and Nashville, Tennessee, according to the U.S. Census Bureau or Tennessee Department of Transportation?'', ``question'': ``How far is Clarksville, TN from Nashville, TN?''\}\} \\
\toolcallclose \\
\toolresponseopen \\
\{``result'': ``[NO\_TARGET\_INFO\_FOUND] The Clarksville, TN--KY metropolitan statistical area consists of Montgomery and Stewart Counties in Tennessee...; The public hearings on April 24, 2012 regarding proposed regional express bus service from Clarksville to Nashville indicate that the distance between Clarksville and Nashville is significant enough to require express bus service.''\} \\
\toolresponseclose \\
The document mentions that public hearings discussed proposed regional express bus service from Clarksville to Nashville, noting the distance is significant. However, it does not provide the exact numerical distance or any quantitative measure (e.g., miles or kilometers), only qualitative information about its significance. \\
\suba{2}: Approximately 95 miles \\
\finalansweropen \\
95 miles \\
\finalanswerclose \\
\bottomrule
\end{tabular}
\caption{Case study illustrating E2.2 error of \textsc{D$^2$Plan}.}
\label{tab:case_e22}
\end{table*}

\section{Prompts}
\label{app:prompts}

\begin{table*}[h!]
\centering
\small
\begin{tabular}{|p{0.95\linewidth}|}
\hline
\rowcolor[gray]{0.4}
\textcolor{white}{\textbf{Question Decomposition Prompt}} \\
\hline
I will give you a multi-hop question. You need to decompose it into atomic sub-questions. Since the sub-questions may depend on each other, you need to number the sub-questions (\#Q\_1, \#Q\_2, \#Q\_3...). A sub-question numbered k can only depend on all sub-questions numbered less than k. When considering dependencies, some sub-questions may rely on the answers to other sub-questions. In this case, you need to denote these answers as \#A\_i, where i is the corresponding sub-question number. \\
\\
Please follow these instructions. \\
- Be creative and varied in your response. \\
- You should output in the \texttt{<output></output>} tags \\
- First provide the rationale for the decomposition, then list the sub-questions. \\
- The reasonable number of sub-questions is between 2 and 5. \\
- Do not output subtitles like ``Rationale for the Decomposition:'', but directly provide the rationale. \\
- You need to place the subquestions within the \texttt{<plan></plan>} tag. \\
\\
An Example of a decomposition: \\
\\
\texttt{<question>} \\
Here is a multi-hop question \\
\texttt{</question>} \\
\\
\texttt{<output>} \\
Here is the rationale for the decomposition \\
\\
\texttt{<plan>} \\
\#Q\_1: Sub-question 1 \\
\#Q\_2: Sub-question 2 \\
... \\
\texttt{</plan>} \\
\texttt{</output>} \\
\\
\textbf{Question:} \{question\} \\
\hline
\end{tabular}
\caption{Prompt template for question decomposition in multi-hop QA.}
\label{tab:decompose_prompt}
\end{table*}

\begin{table*}[h!]
\centering
\small
\begin{tabular}{|p{0.95\linewidth}|}
\hline
\rowcolor[gray]{0.4}
\textcolor{white}{\textbf{Query Rewrite Prompt}} \\
\hline
You are an expert search strategist for a semantic retrieval system. Your goal is to analyze and refine a search query that has previously failed to retrieve relevant documents. \\
\\
Some common rewriting strategies include: expanding the query into a hypothetical answer, breaking down the question into sub-questions, adding or correcting key entities, asking a more general high-level question, or narrowing the query with specific constraints. You may use these or any other strategy you find appropriate. \\
\\
Based on the user's question and the failed search history, your task is to generate a new, improved query. Provide your output as a single JSON object with no other text before or after it. \\
\\
Original question: \{question\} \\
\\
Previous search attempts (all failed to retrieve relevant information): \{query\_history\} \\
\\
JSON Output: \\
You must output a JSON formatted object that includes: \\
- ``reason'': A concise analysis of why the previous queries likely failed and a justification for your new approach. \\
- ``new\_query'': The modified search query, designed for better semantic retrieval. \\
\hline
\end{tabular}
\caption{Prompt template for query rewriting in iterative retrieval.}
\label{tab:modify_query_prompt}
\end{table*}

\begin{table*}[h!]
\centering
\small
\begin{tabular}{|p{0.95\linewidth}|}
\hline
\rowcolor[gray]{0.4}
\textcolor{white}{\textbf{Purifier Agent Prompt}} \\
\hline
TASK: Analyze the provided [DOCUMENTS] to determine whether they contain information that answers the [QUESTION]. Then output EXACTLY ONE valid JSON object following the [RULES] and [JSON OUTPUT FORMAT]. \\
\\
RULES: \\
\\
1. RELEVANT DETERMINATION: \\
\quad If ANY document provides information that can help to answer the QUESTION, set ``relevant'' to ``Yes''; otherwise ``No''. \\
\\
2. WHEN relevant = ``Yes'': \\
\quad - ``extracted\_info'': Output ONLY the minimal factual statements (facts) present in the documents that are strictly required to answer the QUESTION. Do NOT output a full formulated answer, do NOT repeat or restate the QUESTION. \\
\quad - Each fact must be directly supported by the documents; paraphrasing is allowed but no speculation or inference beyond what is stated. \\
\quad - Each fact MUST be SELF-CONTAINED: explicitly name the subject/entity; do NOT use context-dependent pronouns or vague referents (e.g. it, they, he, she, this, that, these, those, its, their, former, latter). Replace such pronouns with the concrete entity name from the documents. \\
\quad - If multiple facts are needed, separate them with ``; '' (semicolon + space). If one fact suffices, output just that fact. \\
\quad - No document indices, no citations, no commentary, no prefixes like ``Fact:'' or ``According to''. \\
\quad - ``summary'': must be an empty string ``''. \\
\\
3. WHEN relevant = ``No'': \\
\quad - ``extracted\_info'': must be ``''. \\
\quad - ``summary'': Provide brief summaries for EACH document using the exact format: \\
\quad\quad {[}Doc 1{]}: <summary of document 1>\textbackslash n{[}Doc 2{]}: <summary of document 2>\textbackslash n... \\
\quad Each summary should be concise and factual. \\
\\
4. OUTPUT: \\
\quad - Return ONLY the JSON object; no extra text, no markdown. \\
\quad - Ensure valid JSON (double quotes, proper escaping). \\
\\
QUESTION: \{query\} \\
\\
DOCUMENTS: \\
\{docs\_text\} \\
\\
JSON OUTPUT FORMAT (fields and meaning): \\
``relevant'': ``Yes'' or ``No'' \\
``extracted\_info'': See RULES \\
``summary'': See RULES \\
\\
OUTPUT: \\
\hline
\end{tabular}
\caption{Prompt template for the Purifier Agent to extract relevant information from retrieved documents.}
\label{tab:purifier_prompt}
\end{table*}

\begin{table*}[h!]
\centering
\small
\begin{tabular}{|p{0.95\linewidth}|}
\hline
\rowcolor[gray]{0.4}
\textcolor{white}{\textbf{Sub-question Answering Prompt}} \\
\hline
You need to answer the question. Please concisely analyze it first before providing your answer. You should provide your answer at the end in the \texttt{<answer></answer>} tag. \\
\\
Rules: \\
- If you think that there is no information in the document to answer the question, please first indicate that there is no information, and then use your internal knowledge to answer the question. \\
- Do NOT include any markdown formatting in your response like ``\# Answer'' or ``\#\# Explanation''. \\
- The final answer in the \texttt{<answer></answer>} tag should be concise (no more than 15 words) and directly address the original question. It can be a single word or a phrase without markdown formatting. \\
\\
\texttt{<document>} \\
\{document\} \\
\texttt{</document>} \\
\texttt{<question>} \\
\{question\} \\
\texttt{</question>} \\
\hline
\end{tabular}
\caption{Prompt template for answering sub-questions with retrieved documents.}
\label{tab:answer_subq_prompt}
\end{table*}

\begin{table*}[h!]
\centering
\small
\begin{tabular}{|p{0.95\linewidth}|}
\hline
\rowcolor[gray]{0.4}
\textcolor{white}{\textbf{Sub-question Update Prompt}} \\
\hline
I will give you a multi-hop question and several sub-questions decomposed from it. Sub-questions 1$\sim$\{k\} have been solved. Now please update sub-question \{j\} based on the questions and answers of sub-questions 1$\sim$\{k\}, making it self-contained. Only output the updated sub-question. \\
\\
\textbf{Multi-hop question:} \\
\{question\} \\
\\
\textbf{Sub-questions and answers:} \\
\{sub\_qa\} \\
\\
\textbf{Updated \#Q\_\{j\}:} \\
\hline
\end{tabular}
\caption{Prompt template for updating sub-questions based on solved dependencies.}
\label{tab:update_subq_prompt}
\end{table*}

\begin{table*}[h!]
\centering
\small
\begin{tabular}{|p{0.95\linewidth}|}
\hline
\rowcolor[gray]{0.4}
\textcolor{white}{\textbf{Retrieval Necessity Judgment Prompt}} \\
\hline
I will give you the original question, previously solved sub-questions and their answers, and the current sub-question to be solved. Please determine whether you need to retrieve external information to answer the current sub-question based on the information already available. \\
\\
You need to output a JSON formatted object that includes: \\
- ``reason'': a concise reason for whether you need to retrieve external information \\
- ``need\_retrieval'': true/false, indicating whether retrieval is needed \\
- ``query'': if need\_retrieval is true, please provide the query for retrieval; otherwise, set to null \\
\\
\textbf{Original question:} \{original\_question\} \\
\\
\textbf{Previously solved sub-questions and answers:} \\
\{previous\_qas\} \\
\\
\textbf{Current sub-question to solve:} \{current\_subquestion\} \\
\hline
\end{tabular}
\caption{Prompt template for judging whether retrieval is needed for a sub-question.}
\label{tab:judge_need_retrieval_prompt}
\end{table*}

\begin{table*}[h!]
\centering
\small
\begin{tabular}{|p{0.95\linewidth}|}
\hline
\rowcolor[gray]{0.4}
\textcolor{white}{\textbf{Plan Revision Prompt}} \\
\hline
I will give you the original multi-hop question and the solved sub-questions with their answers so far. You need to REVISE the remaining plan by generating a new ordered list of self-contained replacement sub-questions that efficiently lead to the final answer. \\
\\
Please follow these instructions. \\
- Do not repeat already-solved sub-questions. \\
- Each new sub-question must be self-contained and should implicitly leverage the known answers. \\
- Keep the list concise (between 1 and 5 items). \\
- A sub-question numbered k can only depend on all sub-questions numbered less than k. When considering dependencies, some sub-questions may rely on the answers to other sub-questions. In this case, you need to denote these answers as \#A\_i, where i is the corresponding sub-question number. \\
- Output strictly within the \texttt{<output></output>} tags. \\
- Inside \texttt{<output>}, write only lines in the following format, one per line: \\
\quad \#Q\_1: <sub-question> \\
\quad \#Q\_2: <sub-question> \\
\quad ... \\
\\
An Example: \\
\\
\texttt{<output>} \\
\#Q\_1: xxxx \\
\#Q\_2: xxxx based on \#A\_1 \\
\#Q\_3: xxxx based on \#A\_1 and \#A\_2 \\
\texttt{</output>} \\
\\
\textbf{Original multi-hop question:} \\
\{question\} \\
\\
\textbf{Solved sub-questions and their answers so far:} \\
\{solved\_qas\} \\
\hline
\end{tabular}
\caption{Prompt template for plan revision after retrieval failures.}
\label{tab:replan_subquestions_prompt}
\end{table*}

\begin{table*}[h!]
\centering
\small
\begin{tabular}{|p{0.95\linewidth}|}
\hline
\rowcolor[gray]{0.4}
\textcolor{white}{\textbf{Final Answer Synthesis Prompt}} \\
\hline
Based on the analysis of the multi-hop question and the sub-questions, please provide a final answer that synthesizes the information from all relevant sub-questions. Only output the final answer. The final answer should be concise (no more than 10 words) and directly address the original multi-hop question. It can be a single word or a phrase without markdown formatting. \\
\\
\textbf{Multi-hop question:} \\
\{question\} \\
\\
\textbf{Sub-questions and answers:} \\
\{sub\_qa\} \\
\\
\textbf{Final answer:} \\
\hline
\end{tabular}
\caption{Prompt template for synthesizing the final answer from sub-question answers.}
\label{tab:conclude_prompt}
\end{table*}

\begin{table*}[h!]
\centering
\begin{tabular}{|p{0.95\linewidth}|}
\hline
\rowcolor[gray]{0.4}
\textcolor{white}{\textbf{LLM-as-a-Judge Prompt Template}} \\
\hline
You are a judge. You need to determine whether the answer to a question provided by the LLM is correct based on the reference answer. Please answer ``Yes'' if they are consistent and ``No'' if they are not. Only output ``Yes'' or ``No''. \\
\\
\textbf{Question:} \{question\} \\
\textbf{LLM answer:} \{llm\_answer\} \\
\textbf{Reference answer:} \{reference\_answer\} \\
\hline
\end{tabular}
\caption{Prompt template for LLM-as-a-Judge (LasJ) evaluation.}
\label{tab:lasj_prompt}
\end{table*}

\begin{table*}[h!]
\centering
\begin{tabular}{|p{0.95\linewidth}|}
\hline
\rowcolor[gray]{0.4}
\textcolor{white}{\textbf{Reasoner Agent Prompt (w/o Global Planning)}} \\
\hline
I will give you a question. You can call a retriever to obtain relevant documents to help answer the question, or you can directly use internal knowledge to respond. If you think there is no relevant information in the search results, you can call the retriever again. You can call the retriever multiple times in the multi-turn conversation. You must conduct reasoning inside \texttt{<think>} and \texttt{</think>} first every time you call the retriever. You should think and analyze it first, and then provide the answer at the end in the \texttt{<answer></answer>} tags. The answer in the \texttt{<answer></answer>} tag should be concise (no more than 15 words) and directly address the original question. \\
\\
\textbf{Question:} \{question\} \\
\hline
\end{tabular}
\caption{Prompt template for the Reasoner Agent without global planning mechanism, used in the ablation study (\textit{w/o Global Planning}).}
\label{tab:prompt_wo_global_planning}
\end{table*}

\begin{table*}[h!]
\centering
\small
\begin{tabular}{|p{0.95\linewidth}|}
\hline
\rowcolor[gray]{0.4}
\textcolor{white}{\textbf{Trajectory Rewriting Prompt for $\mathcal{A}_{r}$}} \\
\hline
You are an expert at transforming structured reasoning trajectories into coherent, natural reasoning processes. Your task is to rewrite the given trajectory into a fluent and logical reasoning flow that can serve as a high-quality training example for a Reasoner Agent. \\
\\
\#\# Input Format \\
You will receive a structured reasoning trajectory.
\\
\#\# Output Requirements \\
Transform the trajectory into a natural reasoning process that: \\
1. Maintains all the key reasoning steps and information from the original trajectory \\
2. Uses a coherent narrative flow with smooth transitions between steps \\
3. Preserves the structured tags required for training: \\
\quad - \texttt{<tool\_call>...</tool\_call>} for each retrieval attempt \\
\quad - \texttt{<tool\_response>...</tool\_response>} for each retrieval result \\
\quad - \texttt{<plan>...</plan>} for initial question decomposition \\
\quad - \texttt{<Updated\_\#Q\_i>...</Updated\_\#Q\_i>} for sub-question updates \\
\quad - \texttt{<Replan>...</Replan>} for global replanning \\
\quad - \texttt{<answer>...</answer>} for the final answer \\
4. Removes redundant or verbose content while keeping essential reasoning \\
5. Ensures the reasoning appears natural and self-contained \\
\\
\#\# Guidelines \\
- The rewritten trajectory should read as if a skilled reasoner is thinking through the problem step by step \\
- Integrate the retrieval feedback naturally into the reasoning flow \\
- When sub-questions are updated, explain the reasoning behind the update \\
- When replanning occurs, articulate why the previous approach failed and how the new plan addresses it \\
- Keep the final answer concise and directly addressing the original question \\
\\
\textbf{Original Trajectory:} \\
\{trajectory\} \\
\\
\textbf{Rewritten Trajectory:} \\
\hline
\end{tabular}
\caption{Prompt template for rewriting synthesized trajectories into coherent reasoning processes for Reasoner's SFT training (\S\ref{subsubsec:constructing_sft_data}).}
\label{tab:trajectory_rewrite_prompt}
\end{table*}

\begin{table*}[h!]
\centering
\small
\begin{tabular}{|p{0.95\linewidth}|}
\hline
\rowcolor[gray]{0.4}
\textcolor{white}{\textbf{Error Analysis Prompt}} \\
\hline
You are a professional Search Agent Error Diagnostic Expert. Your task is to analyze the reasoning trajectory of an Agent in Open-domain Multi-hop QA tasks and classify errors based on the criteria provided below. \\
\\
Please strictly follow the ``First Error'' principle: check the reasoning trajectory chronologically from the beginning. Once you identify an error fitting any of the following categories, immediately stop checking subsequent steps and output the result. \\
\\
\#\# Type 1: Erroneous Question Decomposition \\
The Agent makes a mistake during question decomposition, resulting in an incorrect sequence of sub-questions/sub-queries. Examples include: \\
- The generated series of sub-questions deviates from the original question or covers insufficient knowledge to effectively support the answer to the complex task. \\
- The decomposed sub-questions are not atomic enough or are too granular. \\
- When facing a factual sub-question, the Agent fails to invoke a search tool and directly answers incorrectly based on internal parametric knowledge. \\
\\
\#\# Type 2: Reasoning Hijacking by Peripheral Evidence (Direct Reasoning) \\
The model directly reasons with irrelevant retrieval results without attempting remediation. Specifically, the planning logic and the sub-question itself are correct, but the retrieval results consist entirely of irrelevant information. The Agent makes no attempt to remedy the situation by modifying the Query or re-decomposing the sub-questions; instead, it answers directly based on the irrelevant information or incorrect internal knowledge. \\
\\
\#\# Type 3: Reasoning Hijacking by Peripheral Evidence (Failed Remediation) \\
The model recognizes that the initial retrieval is irrelevant and attempts query rewriting several times but still fails to obtain correct evidence, resulting in an incorrect answer. Specifically, the planning logic and the sub-question itself are correct. The Agent initiates a search, but the retrieval results consist entirely of irrelevant information. The Agent attempts remedial measures (e.g., rewriting the Query, modifying the sub-question decomposition), but the retrieval results still do not contain relevant information. Consequently, the Agent answers incorrectly due to a lack of information or is forced to answer ``not found.'' \\
\\
\#\# Type 4: Other Errors \\
Errors that do not belong to any of the above categories. \\
\\
\# Input Data \\
{[}Original Question{]} \\
\{question\} \\
{[}Reasoning Trajectory{]} \\
\{trajectory\} \\
{[}Golden Answer{]} \\
\{golden\_answer\} \\
\\
\# Output Format \\
Please output the analysis result in JSON format: \\
\texttt{\{} \\
\texttt{\ \ "analysis": "A brief analysis of the reasoning trajectory, explaining why it is classified into this error type.",} \\
\texttt{\ \ "error\_type": "Type 1 / Type 2 / Type 3 / Type 4",} \\
\texttt{\}} \\
\hline
\end{tabular}
\caption{Prompt template for error analysis using \texttt{gemini-3-pro-preview}.}
\label{tab:error_analysis_prompt}
\end{table*}

\begin{table*}[h!]
\centering
\small
\begin{tabular}{|p{0.95\linewidth}|}
\hline
\rowcolor[gray]{0.4}
\textcolor{white}{\textbf{Redundancy Check Prompt}} \\
\hline
You are a question decomposition expert. Please determine whether the model-generated question decomposition plan contains \textbf{redundant subquestions}. \\
\\
\textbf{Definition of Redundant Subquestions}: \\
- The subquestion does \textbf{NOT help} solve the original question; it is unnecessary \\
- The subquestion has \textbf{NO semantic correspondence} in the Gold-annotated question decomposition \\
- Even if the wording differs, as long as they ask about the same thing, they are considered semantically corresponding \\
\\
\textbf{Example of Redundant Question}: \\
- Original question: ``Where is Ulrich Walter's employer headquartered?'' \\
- Gold decomposition: Q1: Ulrich Walter $>>$ employer, Q2: \#1 $>>$ headquarters location (2 steps) \\
- Model decomposition: Q1: Who is Ulrich Walter?, Q2: Who is Ulrich Walter's employer?, Q3: Where is \#A\_2 headquartered? (3 steps) \\
- ``Who is Ulrich Walter?'' is \textbf{redundant} because it does not help solve the original question, and there is no corresponding question in Gold \\
\\
\textbf{Example of Non-Redundant Question}: \\
- Original question: ``Who is the spouse of the Green performer?'' \\
- Gold decomposition: Q1: Green $>>$ performer, Q2: \#1 $>>$ spouse (2 steps) \\
- Model decomposition: Q1: Who is the Green performer?, Q2: Who is the spouse of \#A\_1? (2 steps) \\
- ``Who is the Green performer?'' is \textbf{NOT redundant} because it is equivalent to Gold's Q1 ``Green $>>$ performer'' and is a necessary step to solve the original question \\
\\
\textbf{Notes}: \\
- ``\#1'' in Gold refers to the answer of the first question; ``\#A\_1'' in the model means the same thing \\
- ``X $>>$ Y'' format means asking for the Y attribute of X, equivalent to ``What/Who is the Y of X?'' \\
- As long as the model's subquestion \textbf{semantically corresponds} to a question in Gold, it is NOT redundant \\
- If the model's subquestion has no semantic correspondence in Gold AND does not help solve the original question, it is redundant \\
\\
\textbf{Original Question}: \{original\_question\} \\
\\
\textbf{Gold Decomposition} (\{gold\_steps\} steps): \\
\{gold\_str\} \\
\\
\textbf{Model Decomposition Plan} (\{model\_steps\} steps): \\
\{model\_str\} \\
\\
Please carefully analyze each subquestion from the model: \\
1. Does this subquestion have a semantically corresponding question in Gold? \\
2. If not, does this subquestion help solve the original question? \\
3. If there is no correspondence AND it does not help, then the question is redundant \\
\\
Please respond in the following JSON format (field names must match exactly): \\
\texttt{\{} \\
\texttt{\ \ "has\_extra": true,} \\
\texttt{\ \ "extra\_questions": ["Q1: Who is Ulrich Walter?"],} \\
\texttt{\ \ "explanation": "Q1 does not help solve the original question and has no correspondence in Gold"} \\
\texttt{\}} \\
Or if there are no redundant questions: \\
\texttt{\{} \\
\texttt{\ \ "has\_extra": false,} \\
\texttt{\ \ "extra\_questions": [],} \\
\texttt{\ \ "explanation": "All questions have correspondence or help solve the original question"} \\
\texttt{\}} \\
\\
Output only JSON, nothing else. \\
\hline
\end{tabular}
\caption{Prompt for detecting redundant sub-questions using \texttt{gemini-3-flash-preview}.}
\label{tab:redundancy_check_prompt}
\end{table*}

\end{document}